\documentclass[11pt]{article}
\ifdefined\pdfminorversion\pdfminorversion=7\fi
\usepackage[preprint]{acl}
\usepackage{times}
\usepackage{latexsym}
\usepackage[T1]{fontenc}
\usepackage[utf8]{inputenc}
\usepackage{microtype}
\usepackage{inconsolata}
\usepackage{booktabs}
\usepackage{float}
\usepackage{listings}
\usepackage{textcomp}
\usepackage{array}
\usepackage{multirow}
\usepackage{enumitem}
\setlist[enumerate]{topsep=3pt,itemsep=2pt,parsep=0pt,partopsep=0pt}
\usepackage{graphicx}
\usepackage{placeins}
\usepackage{stfloats}
\usepackage{cuted}
\usepackage{xcolor}
\usepackage{subcaption}
\usepackage{amsmath}
\usepackage{amssymb}
\usepackage{url}

\renewcommand{\small}{\fontsize{10}{12}\selectfont}
\renewcommand{\footnotesize}{\fontsize{10}{12}\selectfont}
\renewcommand{\Large}{\fontsize{14}{16}\selectfont}

\hypersetup{pdftitle={Do Agents Need a New Kind of Paper? SPRA: Structured Paper Representation for Agents},pdfauthor={Yutong Leng, Haojiong Chen, Fukai Shang, Xiaoyang Wang, Mindong Liu, Runyuan Ma, Jiantao Qiu},pdfsubject={Structured scientific paper representations and resource access},pdfkeywords={SPRA, scientific agents, paper representation, Sciverse}}

\newcolumntype{L}[1]{>{\raggedright\arraybackslash}p{#1}}
\newcolumntype{C}[1]{>{\centering\arraybackslash}p{#1}}

\newcommand{\thinrowbrace}{\scalebox{0.65}[3.2]{$\{$}}
\title{Do Agents Need a New Kind of Paper?\\SPRA: Structured Paper Representation for Agents}
\author{Yutong Leng$^{1,2}$ \quad Haojiong Chen$^1$ \quad Fukai Shang$^1$ \\
\textbf{Xiaoyang Wang}$^1$ \quad \textbf{Mindong Liu}$^1$ \quad \textbf{Runyuan Ma}$^1$ \quad \textbf{Jiantao Qiu}$^{1,*}$ \\
{\normalsize $^1$Shanghai Artificial Intelligence Laboratory} \\
{\normalsize $^2$Shanghai Jiao Tong University} \\
{\normalfont\footnotesize \textsuperscript{*}\,Corresponding author.} \\[3pt]
{\normalfont\footnotesize \href{https://sciverse.opendatalab.com/}{Website}\quad\textperiodcentered\quad \href{https://sciverse.opendatalab.com/docs/sciverse/api/paper-schema}{Paper Schema API}\quad\textperiodcentered\quad \href{https://sciverse.opendatalab.com/docs/sciverse/skills/paper-schema}{Agent Skill}\quad\textperiodcentered\quad \href{https://github.com/gavinwxy/ai-paper-structured-extraction}{Extraction Code}}
}

\newcommand{\SPRAExpansionScale}{5}
\newcommand{\SPRAExpansionNeighborCount}{932}

\newcommand{\SPRAIntegrationSeed}{1.779}
\newcommand{\SPRAIntegrationExpanded}{2.032}
\newcommand{\SPRAIntegrationDelta}{\ensuremath{+0.253}}
\newcommand{\SPRAIntegrationCI}{\ensuremath{[+0.195,+0.308]}}
\newcommand{\SPRANoveltySeed}{2.100}
\newcommand{\SPRANoveltyExpanded}{2.172}
\newcommand{\SPRANoveltyDelta}{\ensuremath{+0.072}}
\newcommand{\SPRANoveltyCI}{\ensuremath{[+0.037,+0.105]}}
\newcommand{\SPRAGroundingSeed}{1.848}
\newcommand{\SPRAGroundingExpanded}{1.931}
\newcommand{\SPRAGroundingDelta}{\ensuremath{+0.083}}
\newcommand{\SPRAGroundingCI}{\ensuremath{[+0.029,+0.139]}}
\newcommand{\SPRAQualificationSeed}{2.750}
\newcommand{\SPRAQualificationExpanded}{2.744}
\newcommand{\SPRAQualificationDelta}{\ensuremath{-0.006}}
\newcommand{\SPRAQualificationCI}{\ensuremath{[-0.076,+0.055]}}
\newcommand{\SPRAOverallSeed}{2.302}
\newcommand{\SPRAOverallExpanded}{2.379}
\newcommand{\SPRACompositeHeldDelta}{\ensuremath{+0.100}}
\newcommand{\SPRACompositeHeldCI}{\ensuremath{[+0.061,+0.141]}}
\newcommand{\SPRAIntegrationHeldDelta}{\ensuremath{+0.254}}
\newcommand{\SPRAIntegrationHeldCI}{\ensuremath{[+0.198,+0.310]}}

\begin{document}
\raggedbottom
\maketitle

\begin{figure*}[!b]
\centering
\captionsetup{hypcap=false}
\includegraphics[width=0.98\textwidth,height=0.32\textheight,keepaspectratio]{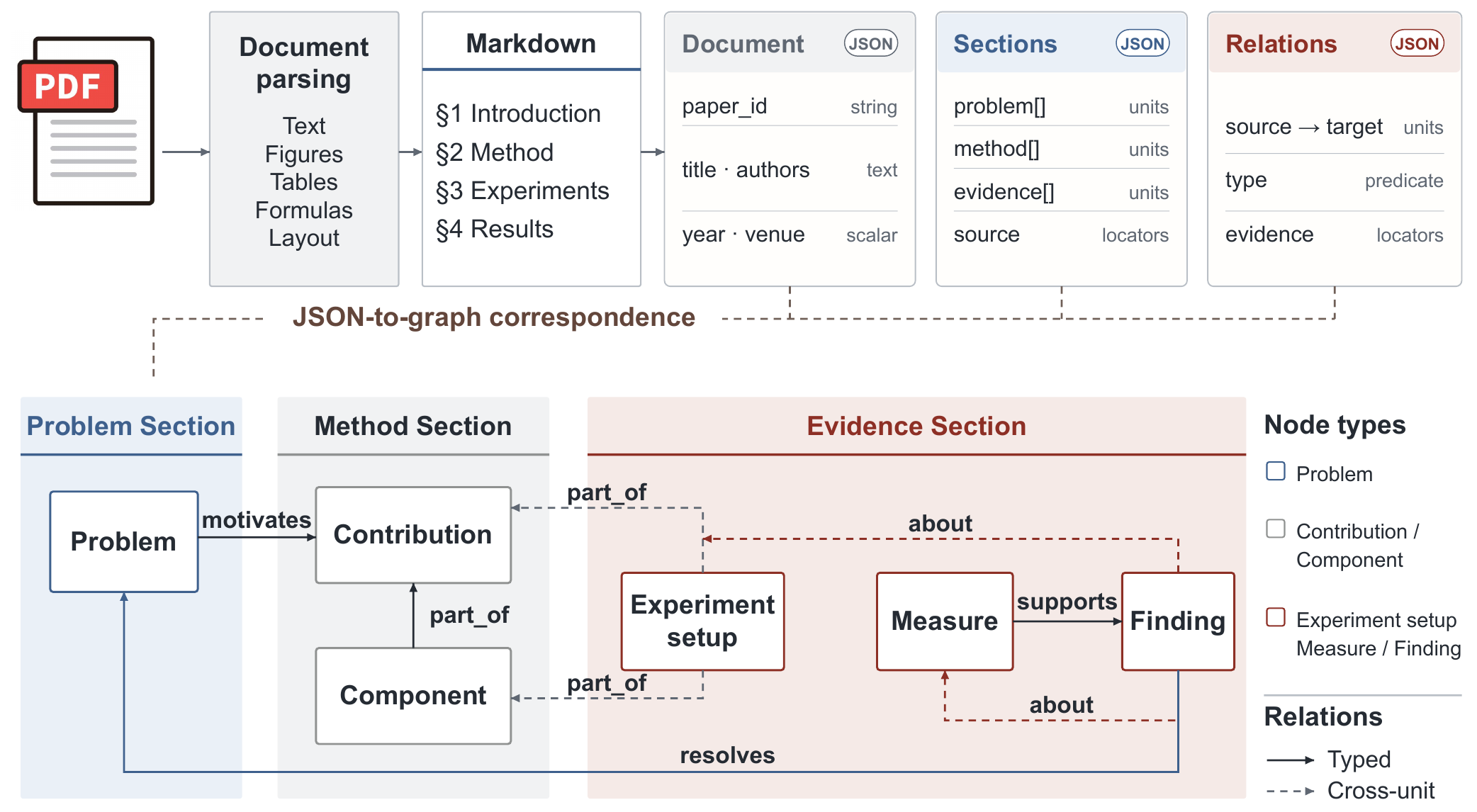}
\caption{Research objects and typed relations in the empirical-AI profile. Problem, Method, and Evidence sections encode the paper's research argument; arrows show relation directions. Units and relations carry source bindings for verification.}
\label{fig:schema}
\end{figure*}

\begin{abstract}
Scientific agents must connect research problems, methods, and findings dispersed through narrative articles. We introduce SPRA (Structured Paper Representation for Agents), a paper-level representation of persistent research objects and typed relations. Its empirical-AI profile organizes Problems, Contributions, Components, ExperimentSetups, Measures, and Findings for querying, traversal, and reuse. Selected paper-level citation relations record dependencies, component use, and comparisons with prior work; source locations support verification. We instantiate SPRA across more than one million AI papers, with 19.95 million research units and 24.07 million internal relations. Evaluations access SPRA through Schema-RAG with task evidence fixed before inference.
\newpage
Across seven subfields, SPRA improves source-conditioned scientific QA by $0.381/5$ over passage retrieval while remaining close to Full Compact ($0.078/5$ lower on average). In directed cross-paper reasoning, qualification improves by $0.184/5$ over passage retrieval; Top-5 synthesis shows mixed quality effects. Across all paired RQ2 tasks, including Top-5, input+output tokens decrease by $28.5\%$ relative to passage retrieval and $95.3\%$ relative to Full Compact. In idea generation, a $0.067$-point lower Overall score than Full Compact accompanies $93.7\%$ fewer input+output tokens. Together, these results support SPRA as a usable representation of scientific content and a foundation for research workflows that connect, inspect, and reuse paper-level research objects. The SPRA resource is accessible through the \href{https://sciverse.opendatalab.com/}{Sciverse} scholarly-data platform.
\end{abstract}
\clearpage

\section{Introduction}
\label{sec:intro}

AI paper submissions are growing rapidly \citep{iclr2026factsheet}, and scientific agents increasingly participate in literature discovery, comparison, and synthesis \citep{he2025pasa,asai2026openscholar}. The same paper can inform method comparisons, evaluation queries, and research proposals. These operations require its problems, contributions, evaluations, and findings, together with their relationships. Conventional articles express this structure through narrative text, tables, and figures; agents must connect findings to methods, conditions, and comparisons. How should a paper expose these scientific relationships for agents to query, compare, and reuse?

Passage-based retrieval supplies relevant source text \citep{lewis2020rag}, while task-specific synthesis assembles evidence for a question \citep{dasigi2021qasper,lala2023paperqa,skarlinski2024paperqa2}. Research roles and dependencies may still need to be reconstructed from these passages. Repeated full-paper inputs incur token-based API charges \citep{chen2023frugalgpt} and occupy context needed for additional papers and task history. Larger context windows also need not ensure reliable evidence use: middle-position information can be underused, with positional attention bias implicated in this behavior \citep{liu2024lost,hsieh2024found}. These costs motivate evaluating retained scientific utility jointly with token use. Compact, explicit research structure makes scientific connections directly available while preserving capacity for linked evidence.

We introduce \textsc{SPRA}, the \emph{Structured Paper Representation for Agents}. SPRA derives its research roles and predicates from paper understanding, method comparison, and evidence integration, representing the connections among a contribution, its evaluation, and its reported findings. Paper-scoped identifiers let successive operations refer to the same stored objects; internal relations connect contributions to their evaluation and reported outcomes, and a separate citation layer records selected dependencies on prior work. Figure~\ref{fig:schema} illustrates SPRA; Schema-RAG is one access mechanism used to evaluate this representation and resource.

Consider the speech-emotion machine-unlearning paper of \citet{ren2025unlearning}. Its privacy-cost Problem, forget-set-only Contribution, unlearning-accuracy Measure, and baseline-comparison Finding form a research argument. An evaluation question selects the linked Measure and Finding; a method comparison selects corresponding records from another paper. To check the reported advantage, an agent can follow these connections to the evaluation conditions and reopen the associated source evidence.

The contributions are:
\begin{enumerate}
  \item a paper-level representation that combines persistent research objects, typed internal relations, citation dependencies, and source bindings, with an extraction and export workflow (Section~\ref{sec:resource});
  \item a million-paper resource, programmatic access, and aligned representation inputs and frozen evaluation outputs (Table~\ref{tab:resource-stats}); and
  \item paired evaluations of scientific QA, cross-paper reasoning, and idea generation that quantify answer quality and task-level token use (Sections~\ref{sec:eval}--\ref{sec:results}).
\end{enumerate}

\begin{table*}[t]
\centering
\small

\begin{tabular}{@{}L{0.20\textwidth}L{0.39\textwidth}L{0.36\textwidth}@{}}
\toprule
Relation & Scientific role & Source $\rightarrow$ Target \\
\midrule
\texttt{part\_of} & Component or setup membership & $\{\mathrm{Co,S}\}\rightarrow\{\mathrm{C,Co,S}\}$ \\
\texttt{co\_contribution} & Joint contributions & $\mathrm{C}\rightarrow\mathrm{C}$ \\
\texttt{compares\_to} & Comparison of methods or measures & $\{\mathrm{C,Co,S,M}\}\rightarrow\{\mathrm{C,Co,S,M}\}$ \\
\texttt{evaluates} & Measure applied to a method & $\mathrm{M}\rightarrow\{\mathrm{C,Co}\}$ \\
\texttt{about} & Subject of a finding & $\mathrm{F}\rightarrow\{\mathrm{C,Co,S,M}\}$ \\
\texttt{supports} & Provides evidence links from research objects to Findings & $\{\mathrm{M,F,C,Co}\}\rightarrow\mathrm{F}$ \\
\texttt{motivates} & Problem motivating a method or setup & $\mathrm{P}\rightarrow\{\mathrm{C,Co,S}\}$ \\
\texttt{resolves} & Finding addressing a problem & $\mathrm{F}\rightarrow\mathrm{P}$ \\
\bottomrule
\end{tabular}
\caption{Paper-internal relations and admissible endpoint signatures. P = Problem; C = Contribution; Co = Component; S = ExperimentSetup; M = Measure; F = Finding. Braces denote allowed types. Construction stages are given in Appendix~\ref{app:construction-spec}.}
\label{tab:relations}
\end{table*}

\section{Related Work}

\paragraph{Scholarly documents and provenance.} GROBID, S2ORC, and MinerU extract metadata, text, and layout \citep{lopez2009grobid,lo2020s2orc,wang2024mineru}. JATS provides markup; PROV-O models provenance; PROV-CONSTRAINTS defines validity; RO-Crate packages research objects \citep{niso2024jats,lebo2013provo,w3c2013provconstraints,soilandreyes2022rocrate,rocrate2026spec}.

\paragraph{Research roles and scientific extraction.} Argumentative Zoning and CoreSC label scientific text by research function \citep{teufel2009az,liakata2012coresc}. SciERC and DyGIE++ model entities, relations, and events \citep{luan2018scierc,wadden2019dygiepp}; SciREX, NLPContributionGraph, SciER, and SciEvent represent document-level relations, contribution triples, or scientific events \citep{jain2020scirex,dsouza2021semeval,zhang2024scier,dong2025scievent}. SPRA assembles research roles and extracted content into persistent paper-owned objects with typed connections and executable operations.

\paragraph{Structured scholarly knowledge.} ORKG supports open scholarly aggregation through paper-linked contributions, templates, comparisons, APIs, and ASK retrieval \citep{jaradeh2019orkg,orkgtemplates,oelen2024orkgask}. SPRA derives roles and predicates from agent tasks and provides an extraction, modeling, and export workflow for paper-owned research objects. These goals are complementary: property alignment connects task-oriented objects to ORKG aggregation. Appendix~\ref{app:orkg-export} describes this engineering interface.

\paragraph{Structured content for scientific assessment.} SciCore combines full-manuscript review with a judgment of an extracted science core, evaluating stability under content-preserving rewrites alongside discrimination across papers \citep{wang2026scicore}. SciCore motivates an explicit view of scientific content for automated assessment. SPRA develops a persistent paper-level representation for broader research operations: paper-owned units, typed relations, and source bindings support querying and comparison across tasks.

\paragraph{Agent access to scientific literature.} PaperQA/PaperQA2, OpenScholar, and PaSa support literature retrieval and cited synthesis \citep{lala2023paperqa,skarlinski2024paperqa2,asai2026openscholar,he2025pasa}; ResearchAgent iteratively develops research ideas \citep{baek2025researchagent}. GraphRAG builds entity communities for query-focused summarization, and KG$^2$RAG expands and organizes retrieved chunks through graph connections \citep{edge2024graphrag,zhu2025kg2rag}. Intern-Atlas exposes method evolution for graph-guided ideation \citep{wu2026internatlas}. SPRA supplies a persistent paper-level research representation that these workflows can query and compose.

\begin{table*}[t]
\centering
\small

\begingroup
\renewcommand{\arraystretch}{1.1}
\begin{tabular*}{\textwidth}{@{\extracolsep{\fill}}llr@{}}
\toprule
Statistic & Scope & Value \\
\midrule
AI papers & AI literature; 1965--2026 & 1,059,112 \\
Core units & Corpus; 18.84/paper & 19,949,289 \\
Typed relations & Corpus; 22.73/paper & 24,073,381 \\
Evaluation sample & 7 subfields; 2025; 1,407 each & 9,849 \\
Locator references & Evaluation sample & 409,082 \\
\bottomrule
\end{tabular*}
\endgroup
\caption{Corpus scale and evaluation scope.}
\label{tab:resource-stats}
\end{table*}

\section{\textsc{SPRA}: Paper-Level Research Representation}
\label{sec:resource}

\subsection{Representation and Profiles}

The construction input is a source article parsed into section-aware Markdown with source markers. The output is a paper-scoped record of research units, typed relations, citation dependencies, and source bindings. Given a research question, an agent selects and connects these stored objects to assemble its evidence input.

SPRA stores research objects independently of a particular query. Structured JSON encodes each instance; a versioned profile specifies scientific roles, relation signatures, cardinalities, and source-linking rules.

The representation follows the paper's scientific argument: a Problem motivates a Contribution; Components and ExperimentSetups describe what is built and tested; Measures define evaluation targets; and Findings record reported observations. These six roles organize the empirical-AI profile. Normally one focused Problem expresses the central research question, with a second permitted for an independent question. Finding attributes record applicable conditions and qualifications. Table~\ref{tab:relations} specifies the relations connecting these objects and their admissible endpoints; citation roles express external dependencies in a separate layer.

The empirical-AI profile instantiates this contract for method and evidence comparison. Profiles make the representation extensible while retaining explicit role and relation constraints.

For paper $p$, write $S_p=(U_p,R_p,C_p,L_p,V_p)$: typed units, internal relations, citation dependencies, source bindings, and source/profile revisions. $C_p$ joins selected citation roles and verbatim signals to bibliographic metadata by citation key. Under profile $\pi$, an internal edge $(u,r,v)\in R_p$ is valid only if $u,v\in U_p$ and their types satisfy $r$'s permitted signature. For example, \texttt{motivates} maps a Problem to a Contribution, Component, or ExperimentSetup. The \texttt{supports} predicate provides evidence links from research objects to a Finding, enabling agents to retrieve associated content and inspect available source evidence. \emph{Schema validity} checks the relation contract, and \emph{source resolution} checks whether a recorded address reaches a source item.

Citation dependencies express how the focal paper uses prior research. \texttt{builds\_on} identifies methodological dependencies, \texttt{uses\_component} records adopted components, and \texttt{compares\_with} records comparisons with cited work. Citation keys join these roles and their verbatim signals to bibliographic records, enabling dependency queries and attribution checks. These paper-level dependencies connect the focal research argument to the prior work it uses.

\paragraph{Running example.} In the unlearning case, the Measure evaluates the forget-set-only Contribution and the Finding records its comparison with a baseline. Paper ownership assigns the proposed method to the focal paper; citation roles identify external work. An agent can retain these objects for comparison and reopen their source locators to check the evaluation conditions. Appendix~\ref{app:cases} gives the records and source bindings.

\subsection{Corpus and Community Access}

Sciverse is the scholarly-data platform that hosts the SPRA collection and provides web and programmatic access to paper records. SPRA denotes the research representation defined in this paper; Sciverse provides its deployed resource and access services.

The September 2026 SPRA snapshot contains 1,059,112 structured paper records spanning 1965--2026, drawn from AI conference literature and preprints, including arXiv. Each paper is processed through the construction workflow below into a stored SPRA instance, accessible through the structured-paper services described in Section~\ref{sec:public-access}.

The evaluation sample comprises 9,996 candidate papers, of which 9,849 are source-aligned and eligible. This domain-stratified sample covers 2025 publications, with 1,407 eligible papers in each of seven AI subfields (Table~\ref{tab:resource-stats}). Structural checks and task evaluations therefore assess deployed representations from the same construction workflow, with balanced coverage across the selected fields.

The accompanying data archive includes the complete 9,849-paper aligned RQ1 cohort with all three input representations and frozen study outputs; the software archive provides offline reanalysis scripts. Project software is Apache-2.0; benchmark data and documentation are CC BY 4.0, with source-derived content retaining its own terms. Appendix~\ref{app:resource-access} details access and licensing; Section~\ref{sec:public-access} describes how Sciverse exposes SPRA to researchers and agents.

\subsection{Public Interfaces and Data Access}
\label{sec:public-access}

Sciverse exposes structured paper records under the service name \emph{Paper Schema}. The \path{/paper-schema} prefix identifies this service in API requests. SPRA specifies the representation studied here, while Paper Schema names the deployed access interface.

Researchers can explore SPRA through \href{https://sciverse.opendatalab.com/}{Sciverse} and access the same paper-level research objects programmatically through the \href{https://sciverse.opendatalab.com/docs/sciverse/api/paper-schema}{Paper Schema API}. For example, a JSON \texttt{POST} to \path{/paper-schema/search} returns candidate paper identifiers; \path{/paper-schema/materials} then retrieves research objects and relations for selected papers. Appendix~\ref{app:public-quickstart} gives the HTTPS address, token authentication, and cURL and Python calls.

The \href{https://sciverse.opendatalab.com/docs/sciverse/skills/paper-schema}{Paper Schema Agent Skill} makes this access sequence available to agents through task-oriented tools for paper discovery, object retrieval, and evidence assembly. \href{https://github.com/opendatalab/Sciverse-Agent-Tools}{Sciverse Agent Tools} provides the integration resources for using these tools in agent workflows. The SPRA construction procedure and its extraction implementation are described in Section~\ref{sec:construction-implementation}.

The online corpus and the frozen evaluation materials serve complementary uses. Token-authenticated API access supports discovery and retrieval of the evolving paper collection. The accompanying \path{data.zip} and \path{software.zip} preserve the study records and offline reanalysis workflow described in Appendix~\ref{app:reproducibility}. Corpus statistics refer to the September 2026 snapshot; later service responses may reflect updated records. API access follows account permissions and quotas, while redistribution of source-derived content follows its source terms.

\subsection{Construction Implementation}
\label{sec:construction-implementation}

\begin{figure*}[t]
\centering
\includegraphics[width=1.0\textwidth]{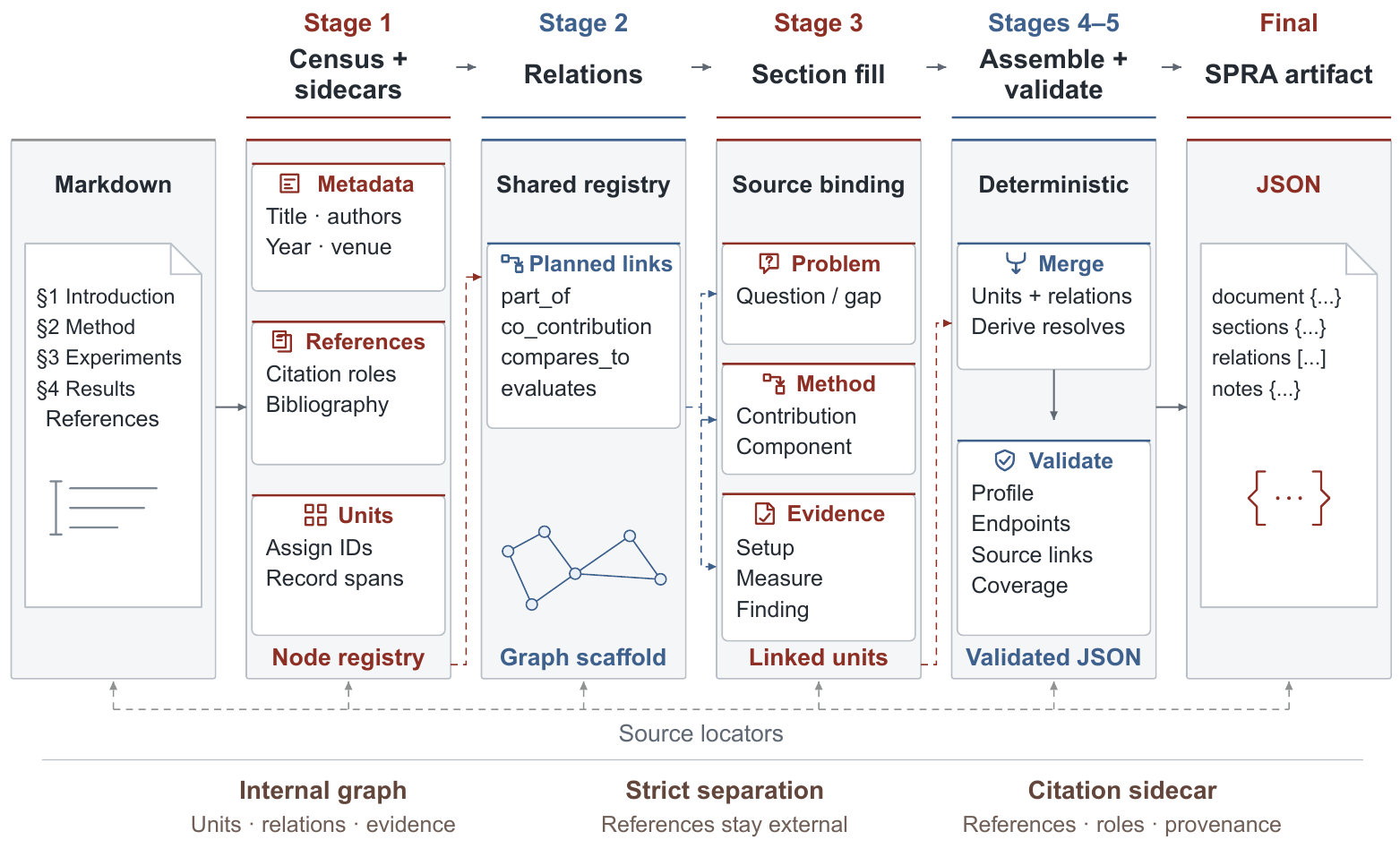}
\caption{Five conceptual stages of SPRA construction. A shared unit census precedes relation planning and role-specific filling; assembly joins source-linked records and stores external citations separately. Validation checks profile conformance, relation endpoints, and source-link integrity.}
\label{fig:pipeline}
\end{figure*}

The \href{https://github.com/gavinwxy/ai-paper-structured-extraction}{SPRA extraction implementation} starts from section-aware Markdown with paragraph markers produced by MinerU (Figure~\ref{fig:pipeline}). Five core LLM calls establish a shared unit census, predict relations, and fill Problem, Method, and Evidence content. Two citation passes join roles and verbatim signals to bibliographic metadata by citation key; implementation and decoding settings appear in Appendix~\ref{app:construction-spec}.

The assembly code resolves source markers, captures tables verbatim, reconciles identifiers, and checks types, endpoints, provenance, and census coverage. Optional numerical checks flag score values absent from captured table cells and explicitly record unchecked cases. These construction controls preserve dense evidence and make numerical transcription inspectable (Appendix~\ref{app:construction-spec}).

\subsection{Operations on Research Objects}

Typed lookup selects methods or findings; relation traversal connects them to measures and settings; faceted filtering selects research roles or attributes. The Sciverse Paper Schema API assembles goal-specific objects, relations, and evidence for paper overviews, literature surveys, benchmark analysis, method comparison, and reproduction. Related-paper discovery extends paper or object queries into literature sets; source opening resolves recorded locators for verification.

Sciverse presents research objects through reading paths and topic graphs, with source addresses for verification; the supplement includes an offline demonstration. The shared registry (Figure~\ref{fig:pipeline}) preserves object identity across extraction passes and later tasks.

\begin{table*}[t]
\centering
\footnotesize

\setlength{\tabcolsep}{3pt}
\begin{tabular}{@{}L{0.14\textwidth}L{0.40\textwidth}L{0.42\textwidth}@{}}
\toprule
Study & Condensed task & Required evidence \\
\midrule
RQ1 G-U3 & Identify the problem and central contribution. & Problem + Contribution \\
RQ1-qual. & Bind qualifications to claims or methods. & Finding + qualification \\
RQ2 & Assess directed relations; synthesize papers. & Typed relations + multi-paper context \\
RQ3 & Generate a grounded benchmark-transfer idea. & Selected papers + qualifications \\
\bottomrule
\end{tabular}
\caption{Representative benchmark operations. Task formulations are condensed; the reproducibility package contains the complete prompts.}
\label{tab:benchmark-operations}
\end{table*}

\section{Evaluating Use of the Representation}
\label{sec:eval}

RQ1--RQ3 assess scientific task quality under fixed research content and a common primary Reader. RQ2 and RQ3 additionally measure input and output token use for the evaluated workflows. Sciverse supplies storage, retrieval, and maintenance; Appendix~\ref{app:cost} defines the token-accounting scope.

\subsection{Representation Quality}

\paragraph{Structural validity.} Structural checks assess recorded source pointers and relation endpoints in the sampled object registries. They measure addressability and registry connectivity.

\subsection{Functional Adequacy of Extracted Content}

RQ1 evaluates source-conditioned scientific QA across problems and contributions, methods and components, experiments and results, and limitations and assumptions. G-U1 measures correctness, coverage, and qualification against shared Full Compact-generated references; G-U3 focuses on Problem + Contribution. Structured questions test research-unit lookup, internal-relation recovery, and qualification filtering. Full Compact supplies complete parsed article text (Appendix~\ref{app:full-compact-input}); model roles appear below.

\subsection{Tasks and Research Questions}

\paragraph{RQ1: source-conditioned scientific QA.} The evaluation frame contains 9,849 eligible papers across seven AI subfields and seven question families for semantic QA and structured operations.

\paragraph{RQ2: cross-paper reasoning.} We evaluate 3,070 complete paired tasks. We score correctness, grounding, and qualification on directed-relation and Top-5 synthesis questions.

\paragraph{RQ3: downstream idea generation.} We evaluate 1,001 task variants across seven AI domains, using shared paper selections to generate and assess evidence-grounded proposals.

\subsection{Controlled Representation Comparison}

Conditions share questions, paper identities and order, answer format, rubric, and Reader. \emph{Full Compact} supplies complete parsed article text; \emph{Text-RAG}, verbatim task-relevant passages; and \emph{Schema-RAG}, SPRA objects and internal relations. Complete multi-paper inputs can exceed context capacity; retrieval keeps evidence compact. Task evidence is fixed before inference. Both RQ2 retrieval conditions select one primary item per paper for each of four shared facets. Text-RAG ranks passages with question-derived BM25F-like queries; Schema-RAG ranks deterministically assembled SPRA objects and typed relations.

\subsection{Reader and Evaluation Protocol}
\label{sec:reader-protocol}

\begin{figure}[!t]
\centering
\includegraphics[width=0.96\columnwidth]{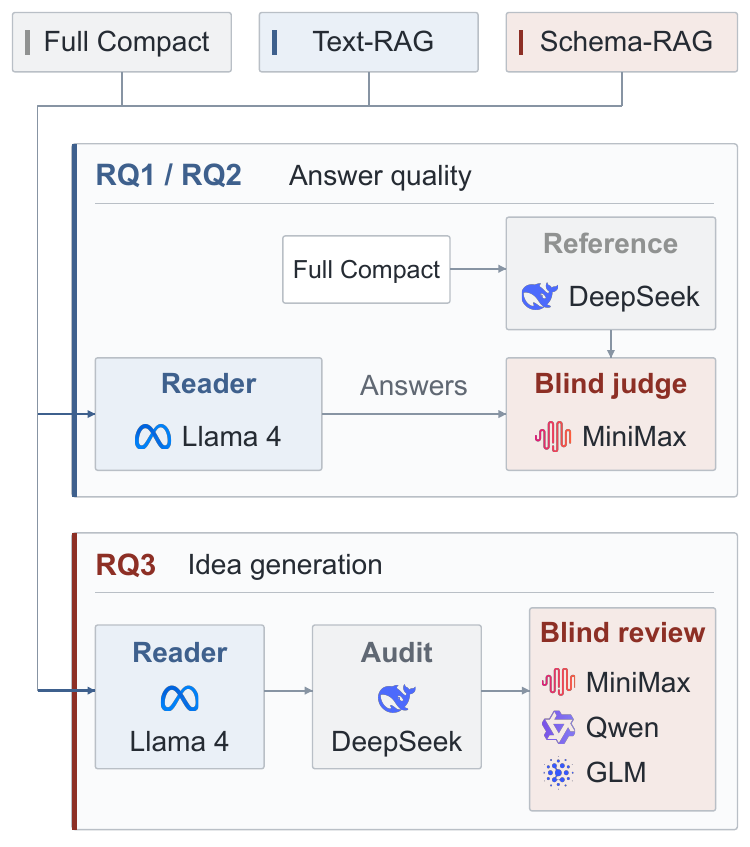}
\caption{Reader and evaluation protocol for the three representation conditions. Section~\ref{sec:reader-protocol} specifies model roles and scoring.}
\label{fig:evaluation-models}
\end{figure}

Figure~\ref{fig:evaluation-models} summarizes model roles and criterion-based evaluation \citep{liu2023geval,zheng2023judge,kim2025biggen}. Llama 4 Scout is the primary Reader; two long-context RQ2 Full Compact tasks use a GPT fallback (Appendix~\ref{app:additional-results}). For RQ1/RQ2, DeepSeek-v4-pro generates shared reference answers from Full Compact sources. MiniMax-M3 scores Reader outputs against these references without condition labels. For RQ3, DeepSeek-v4-pro assesses implementation, and MiniMax-M2.5, Qwen3.5-27B, and GLM-5.1 independently review proposals.

Frozen outputs and scripts support offline replay of RQ1--RQ3 statistics; Appendices~\ref{app:construction-quality} and~\ref{app:cost} document structural checks and token-accounting scope.

\subsection{Metrics and Statistical Analysis}

The primary reported outcomes compare Schema-RAG access with Text-RAG for RQ1 G-U1 and RQ2 directed correctness, and with Full Compact for RQ2 Top-5 correctness. G-U2, G-U3, grounding, qualification, and token use provide supporting analyses. RQ1 and RQ2 quality scores use a 0--5 scale. RQ2 qualification measures the conditions attached to a specified answer. RQ3 reports Overall quality and input+output tokens.

Paired bootstrap resampling preserves condition pairing and study sampling structure to estimate 95\% confidence intervals \citep{koehn2004bootstrap}; the three RQ3 Overall contrasts receive Holm correction \citep{holm1979multiple}. Appendix~\ref{app:additional-results} specifies resampling units, replicate counts, and tests.

\FloatBarrier
\begin{table}[!t]
\centering
\footnotesize

\begin{tabular}{@{}L{0.34\columnwidth}C{0.19\columnwidth}C{0.19\columnwidth}C{0.19\columnwidth}@{}}
\toprule
Measure & Full Compact & Text-RAG & Schema-RAG \\
\midrule
G-U1 & \textbf{2.420} & 1.961 & 2.342 \\
G-U2 & \textbf{1.327} & 1.204 & 1.317 \\
G-U3 & 3.201 & 2.167 & \textbf{3.216} \\
\bottomrule
\end{tabular}
\caption{RQ1 mean scores on the 0--5 scale. G-U1: semantic answer quality judged against references generated from Full Compact; G-U2: structured operations; G-U3: Problem + Contribution. Columns show Full Compact, passage, and SPRA access; bold marks maxima.}
\label{tab:rq1-detail}
\end{table}

\begin{figure*}[!b]
\centering
\includegraphics[width=\textwidth]{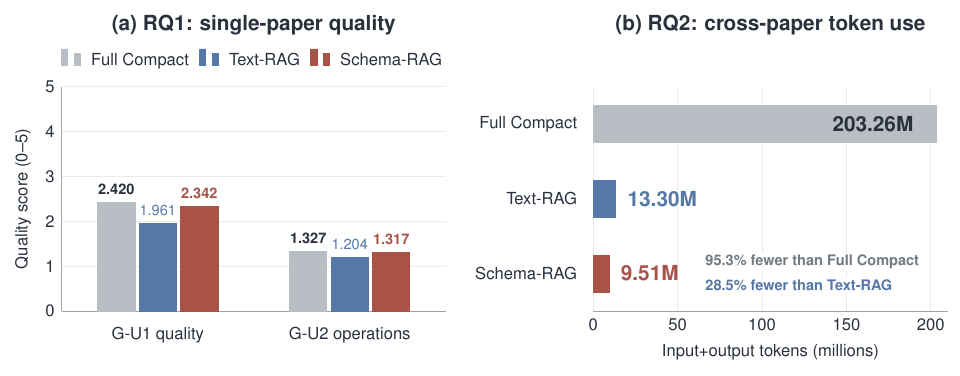}
\caption{Quality and token use: (a) RQ1 G-U1 semantic QA against Full Compact-generated references and G-U2 structured operations (0--5); (b) input+output tokens across all paired RQ2 tasks, including Top-5. Gray and blue annotations show SPRA's reductions relative to Full Compact and Text-RAG.}
\label{fig:rq1-quality}
\end{figure*}

\begin{table*}[t]
\centering
\footnotesize

\setlength{\tabcolsep}{2.4pt}
\renewcommand{\arraystretch}{1.25}
\begin{tabular*}{\textwidth}{@{\extracolsep{\fill}}lclrrrl@{}}
\toprule
Task & & Metric & Full Compact & Text-RAG & Schema-RAG & Difference / reduction \\
\midrule
\multicolumn{7}{@{}l}{\textit{Answer quality (0--5)}} \\
\multirow{3}{*}{Directed} & \multirow{3}{*}{\thinrowbrace} & Correctness & 2.583 & 4.543 & \textbf{4.569} & $+0.026\ [+0.004,+0.049]$ \\
 & & Grounding & 3.098 & 4.323 & \textbf{4.368} & $+0.045\ [+0.016,+0.073]$ \\
 & & Qualification & 2.756 & 4.104 & \textbf{4.288} & $+0.184\ [+0.154,+0.212]$ \\
\addlinespace[1pt]
\multirow{3}{*}{Top-5} & \multirow{3}{*}{\thinrowbrace} & Correctness & \textbf{3.699} & 2.958 & 3.114 & $+0.156\ [+0.114,+0.200]$ \\
 & & Grounding & \textbf{3.439} & 3.110 & 2.966 & $-0.145\ [-0.183,-0.105]$ \\
 & & Qualification & \textbf{3.725} & 2.858 & 2.533 & $-0.325\ [-0.366,-0.284]$ \\
\midrule
\multicolumn{7}{@{}l}{\textit{Input+output (millions of tokens)}} \\
All tasks & & Total tokens & 203.26 & 13.30 & \textbf{9.51} & $-$28.5\% / $-$95.3\% \\
\bottomrule
\end{tabular*}
\caption{RQ2 scores (0--5) and input+output tokens across 3,070 paired tasks. Quality differences are Schema-RAG minus Text-RAG [paired 95\% CI]; token reductions are relative to Text-RAG / Full Compact, respectively. Bold marks the highest quality score in each row and the lowest token total.}
\label{tab:rq2-results}
\end{table*}

\section{Results}
\label{sec:results}

\subsection{Structural and Semantic Assessment}
\label{sec:structural-results}

In the sampled records, 203,628 of 214,239 core units (95.0\%) in the 9,849-paper frame have a source pointer, and 106,439 of 109,591 relations (97.1\%) in the complete relation records for 4,350 papers join to both endpoints. These rates measure addressability and connectivity; RQ1 evaluates content through scientific QA and structured operations. Appendix~\ref{app:construction-quality} gives the structural checks.

In a model-based assessment against public NCG annotations, SPRA receives a mean contribution-level semantic-alignment score of 84.49/100 from GPT-4.1. Information-unit alignment reaches 4.436/5, supporting correspondence between the extracted research roles and annotated contributions (Appendix~\ref{app:ncg-semantic}).

\subsection{Extracted Structure for Source-Conditioned QA}

SPRA retains much of Full Compact's source-conditioned QA score while outperforming passage retrieval (Table~\ref{tab:rq1-detail}; Figure~\ref{fig:rq1-quality}a). G-U1 reaches 2.342/5 against Full Compact's 2.420/5: the paired difference is $-0.078/5$ (95\% CI [$-0.110,-0.046$]). Relative to passage retrieval, the gain is $0.381/5$ (95\% CI [+0.351,+0.411]). G-U1 combines correctness, coverage, and qualification against complete-article references.

For structured operations, SPRA exceeds Text-RAG by 0.113/5 on G-U2. It also has the highest observed G-U3 mean for identifying the problem and contribution, 3.216/5 versus 3.201/5 for Full Compact. These results support the use of SPRA for answering source-conditioned questions and identifying a paper's problem and contribution.

\subsection{Cross-Paper Reasoning}

Across 3,070 paired tasks (Appendix~\ref{app:rq2-token-accounting}), SPRA uses 9.51M input+output tokens: 28.5\% fewer than Text-RAG (13.30M) and 95.3\% fewer than Full Compact (203.26M; Table~\ref{tab:rq2-results}).

Schema-RAG improves directed judgments across all three measured criteria (Table~\ref{tab:rq2-results}). Relative to Text-RAG, qualification improves by 0.184/5 (95\% CI [+0.154,+0.212]), correctness by 0.026/5, and grounding by 0.045/5. The largest gain concerns qualification: retaining support, uncertainty, and applicable conditions.

Top-5 synthesis shows a different result pattern. SPRA improves correctness over Text-RAG by 0.156/5, while grounding and qualification decrease by 0.145/5 and 0.325/5; Full Compact has the highest means on all three criteria. These synthesis results motivate further evaluation of input selection that retains claim-specific qualifications alongside core findings.

\subsection{Idea Generation and Citation Expansion}

The main RQ3 comparison evaluates evidence-grounded proposals across 1,001 variants with shared paper selections. The family-balanced SPRA--Full Compact Overall contrast is $-0.067$ points (95\% CI [$-0.102,-0.034$]); the SPRA--Text-RAG contrast is +0.011 points (95\% CI [$-0.029,+0.058$]). SPRA uses 93.7\% fewer input+output tokens than Full Compact and 17.7\% fewer than Text-RAG. Appendix~\ref{app:additional-results} reports all three paired contrasts and token totals.

The separate citation-expansion study tests access to additional research through represented dependencies. Using Qwen3.5-397B, both conditions start from the same seed objects under a shared 10,000-token ceiling. Adding up to two resolved citation neighbors improves cross-paper integration by \SPRAIntegrationDelta{}/\SPRAExpansionScale{} (95\% CI \SPRAIntegrationCI{}). Table~\ref{tab:relation-expansion-main} reports dimension-specific results. Appendix~\ref{app:relation-expansion} specifies neighbor selection, added evidence, and paired analysis.

\begin{table}[t]
\centering\footnotesize

\begin{tabular}{@{}L{0.40\linewidth}C{0.25\linewidth}C{0.28\linewidth}@{}}
\toprule
Metric & SPRA seeds & SPRA + citations \\
\midrule
Cross-paper integration & \SPRAIntegrationSeed{} & \textbf{\SPRAIntegrationExpanded{}} \\
Novelty & \SPRANoveltySeed{} & \SPRANoveltyExpanded{} \\
Evidence grounding & \SPRAGroundingSeed{} & \SPRAGroundingExpanded{} \\
Qualification handling & \SPRAQualificationSeed{} & \SPRAQualificationExpanded{} \\
Independent Overall & \SPRAOverallSeed{} & \SPRAOverallExpanded{} \\
\bottomrule
\end{tabular}
\caption{Scientific quality with SPRA seeds and citation expansion: 1,001 paired questions; quality scores use 0--5. Means average two reviewers; Overall is assessed independently.}
\label{tab:relation-expansion-main}
\end{table}

\FloatBarrier
\section{Discussion}

SPRA makes the research argument an explicit, reusable object of scientific work. Findings remain connected to contributions, evaluations, and conditions, while selected citation dependencies connect the focal paper to prior work. These connections support queries and relation traversal. Citation expansion demonstrates their use in assembling additional evidence under a shared context ceiling.

The evaluations connect representation quality to scientific use across QA, cross-paper reasoning, and idea generation. Token savings quantify workflow efficiency alongside the scientific utility of explicit research structure.

\section{Conclusion}

SPRA addresses the title's question by exposing the research structure already present in scientific papers as persistent, queryable objects and relations. The empirical-AI profile connects problems, contributions, methods, evaluations, and findings, with source bindings for inspection and selected citation dependencies for cross-paper access. Sciverse makes this representation accessible at million-paper scale through a website, the Paper Schema API, and an Agent Skill; the extraction implementation supports construction from parsed articles.

The evaluations show that SPRA supports source-conditioned QA and directed cross-paper reasoning with substantially lower token use in the evaluated retrieval workflows. RQ1 gains $0.381/5$ over passage retrieval while scoring $0.078/5$ below Full Compact; RQ2 uses $28.5\%$ fewer input+output tokens than passage retrieval across all paired tasks. Mixed Top-5 synthesis results and the lower RQ3 Overall score relative to Full Compact show that compact structure must be assessed against each workflow's information needs. SPRA provides a concrete foundation for research systems that query, compare, and reuse paper-level scientific content through a shared representation.

\section*{Limitations}
\label{sec:limitations}

\paragraph{Generalization and successive workflows.} The evaluations cover the empirical-AI profile and the specified task workflows. Transfer to other domains, successive-task reuse, and stability under meaning-preserving rewrites remain future evaluation directions. The comparisons assess the instantiated representation and access workflows; they do not isolate the contribution of individual role or predicate labels.

\paragraph{Services and cost coverage.} Updates to hosted services can change behavior in new runs; the results characterize the recorded service invocations. Token savings cover model inputs and outputs for downstream tasks; storage, retrieval, and maintenance are provided by Sciverse.

\section*{Ethical Considerations}

SPRA preserves attribution and inspectable source links. Automated extraction may omit or misstate qualifications, which can change how a research claim is understood. Corpus coverage is shaped by the included venues, years, and source availability, so missing or underrepresented work should not be taken as evidence that a topic or finding is absent from the literature. Users may also over-rely on concise structured records and lose context available in the full article. For important research or other consequential uses, users should check the relevant claims and qualifications against the source article. Access and redistribution follow source licenses; researchers retain responsibility for interpretation and downstream validation.

\section*{Acknowledgements}

OpenAI Codex assisted with editorial revision and routine monitoring of experimental progress. The authors reviewed the assisted work and take responsibility for the research decisions, analyses, citations, interpretation, and presentation.

\begingroup
\let\spraBibliography\thebibliography
\renewcommand{\thebibliography}[1]{\spraBibliography{#1}\setlength{\itemsep}{1pt}\setlength{\parsep}{0pt}}
\bibliography{paper_naacl_refs}

\begin{thebibliography}{43}
\providecommand{\natexlab}[1]{#1}

\bibitem[{Asai et~al.(2026)Asai, He, Shao, Shi, Singh, Chang, Lo, Soldaini,
  Feldman, D'Arcy, Wadden, Latzke, Sparks, Hwang, Kishore, Tian, Ji, Liu, Tong,
  Wu, Xiong, Zettlemoyer, Neubig, Weld, Downey, Yih, Koh, and
  Hajishirzi}]{asai2026openscholar}
Akari Asai, Jacqueline He, Rulin Shao, Weijia Shi, Amanpreet Singh, Joseph~Chee
  Chang, Kyle Lo, Luca Soldaini, Sergey Feldman, Mike D'Arcy, David Wadden,
  Matt Latzke, Jenna Sparks, Jena~D. Hwang, Varsha Kishore, Minyang Tian, Pan
  Ji, Shengyan Liu, Hao Tong, and 9 others. 2026.
\newblock \href {https://doi.org/10.1038/s41586-025-10072-4} {Synthesizing
  scientific literature with retrieval-augmented language models}.
\newblock \emph{Nature}, 650(8103):857--863.

\bibitem[{Baek et~al.(2025)Baek, Jauhar, Cucerzan, and
  Hwang}]{baek2025researchagent}
Jinheon Baek, Sujay~Kumar Jauhar, Silviu Cucerzan, and Sung~Ju Hwang. 2025.
\newblock \href {https://doi.org/10.18653/v1/2025.naacl-long.342}
  {{R}esearch{A}gent: Iterative research idea generation over scientific
  literature with large language models}.
\newblock In \emph{Proceedings of the 2025 Conference of the Nations of the
  Americas Chapter of the Association for Computational Linguistics: Human
  Language Technologies (Volume 1: Long Papers)}, pages 6709--6738,
  Albuquerque, New Mexico. Association for Computational Linguistics.

\bibitem[{Brzovic et~al.(2025)Brzovic, Rojas, and Abeliuk}]{brzovic2025entropy}
Sebastián Brzovic, Cristóbal Rojas, and Andrés Abeliuk. 2025.
\newblock \href {https://arxiv.org/abs/2510.11524v1} {Networks multiscale
  entropy analysis}.
\newblock \emph{Preprint}, arXiv:2510.11524v1.

\bibitem[{Chen et~al.(2023)Chen, Zaharia, and Zou}]{chen2023frugalgpt}
Lingjiao Chen, Matei Zaharia, and James Zou. 2023.
\newblock \href {https://arxiv.org/abs/2305.05176} {{FrugalGPT}: How to use
  large language models while reducing cost and improving performance}.
\newblock \emph{Preprint}, arXiv:2305.05176.

\bibitem[{Cheney et~al.(2013)Cheney, Missier, and
  Moreau}]{w3c2013provconstraints}
James Cheney, Paolo Missier, and Luc Moreau. 2013.
\newblock \href {https://www.w3.org/TR/prov-constraints/} {Constraints of the
  {PROV} data model}.
\newblock {W3C} recommendation, World Wide Web Consortium.

\bibitem[{Dasigi et~al.(2021)Dasigi, Lo, Beltagy, Cohan, Smith, and
  Gardner}]{dasigi2021qasper}
Pradeep Dasigi, Kyle Lo, Iz~Beltagy, Arman Cohan, Noah~A. Smith, and Matt
  Gardner. 2021.
\newblock \href {https://doi.org/10.18653/v1/2021.naacl-main.365} {A dataset of
  information-seeking questions and answers anchored in research papers}.
\newblock In \emph{Proceedings of the 2021 Conference of the North American
  Chapter of the Association for Computational Linguistics: Human Language
  Technologies}, pages 4599--4610, Online. Association for Computational
  Linguistics.

\bibitem[{Dong et~al.(2025)Dong, Shah, Sonawane, Banerjee, Brady, Du, and
  Jiang}]{dong2025scievent}
Bofu Dong, Pritesh Shah, Sumedh Sonawane, Tiyasha Banerjee, Erin Brady, Xinya
  Du, and Ming Jiang. 2025.
\newblock \href {https://doi.org/10.18653/v1/2025.emnlp-main.871}
  {{S}ci{E}vent: Benchmarking multi-domain scientific event extraction}.
\newblock In \emph{Proceedings of the 2025 Conference on Empirical Methods in
  Natural Language Processing}, pages 17225--17255, Suzhou, China. Association
  for Computational Linguistics.

\bibitem[{D{'}Souza et~al.(2021)D{'}Souza, Auer, and
  Pedersen}]{dsouza2021semeval}
Jennifer D{'}Souza, S{\"o}ren Auer, and Ted Pedersen. 2021.
\newblock \href {https://doi.org/10.18653/v1/2021.semeval-1.44}
  {{S}em{E}val-2021 task 11: {NLPC}ontribution{G}raph - structuring scholarly
  {NLP} contributions for a research knowledge graph}.
\newblock In \emph{Proceedings of the 15th International Workshop on Semantic
  Evaluation (SemEval-2021)}, pages 364--376, Online. Association for
  Computational Linguistics.

\bibitem[{Edge et~al.(2024)Edge, Trinh, Cheng, Bradley, Chao, Mody, Truitt,
  Metropolitansky, Ness, and Larson}]{edge2024graphrag}
Darren Edge, Ha~Trinh, Newman Cheng, Joshua Bradley, Alex Chao, Apurva Mody,
  Steven Truitt, Dasha Metropolitansky, Robert~Osazuwa Ness, and Jonathan
  Larson. 2024.
\newblock \href {https://arxiv.org/abs/2404.16130} {From local to global: A
  graph rag approach to query-focused summarization}.
\newblock \emph{Preprint}, arXiv:2404.16130.
\newblock Version 2, revised February 2025.

\bibitem[{He et~al.(2025)He, Huang, Feng, Lin, Zhang, Li, and E}]{he2025pasa}
Yichen He, Guanhua Huang, Peiyuan Feng, Yuan Lin, Yuchen Zhang, Hang Li, and
  Weinan E. 2025.
\newblock \href {https://doi.org/10.18653/v1/2025.acl-long.572} {{P}a{S}a: An
  {LLM} agent for comprehensive academic paper search}.
\newblock In \emph{Proceedings of the 63rd Annual Meeting of the Association
  for Computational Linguistics (Volume 1: Long Papers)}, pages 11663--11679,
  Vienna, Austria. Association for Computational Linguistics.

\bibitem[{Holm(1979)}]{holm1979multiple}
Sture Holm. 1979.
\newblock \href {https://www.jstor.org/stable/4615733} {A simple sequentially
  rejective multiple test procedure}.
\newblock \emph{Scandinavian Journal of Statistics}, 6(2):65--70.

\bibitem[{Hsieh et~al.(2024)Hsieh, Chuang, Li, Wang, Le, Kumar, Glass, Ratner,
  Lee, Krishna, and Pfister}]{hsieh2024found}
Cheng-Yu Hsieh, Yung-Sung Chuang, Chun-Liang Li, Zifeng Wang, Long Le, Abhishek
  Kumar, James Glass, Alexander Ratner, Chen-Yu Lee, Ranjay Krishna, and Tomas
  Pfister. 2024.
\newblock \href {https://doi.org/10.18653/v1/2024.findings-acl.890} {Found in
  the middle: Calibrating positional attention bias improves long context
  utilization}.
\newblock In \emph{Findings of the Association for Computational Linguistics:
  ACL 2024}, pages 14982--14995. Association for Computational Linguistics.

\bibitem[{{ICLR}(2026)}]{iclr2026factsheet}
{ICLR}. 2026.
\newblock \href
  {https://media.iclr.cc/Conferences/ICLR2026/ICLR2026_Fact_Sheet.pdf} {{ICLR}
  2026 fact sheet}.
\newblock International Conference on Learning Representations.
\newblock Accessed 2026-10-13.

\bibitem[{Jain et~al.(2020)Jain, van Zuylen, Hajishirzi, and
  Beltagy}]{jain2020scirex}
Sarthak Jain, Madeleine van Zuylen, Hannaneh Hajishirzi, and Iz~Beltagy. 2020.
\newblock \href {https://doi.org/10.18653/v1/2020.acl-main.670} {{S}ci{REX}:
  {A} challenge dataset for document-level information extraction}.
\newblock In \emph{Proceedings of the 58th Annual Meeting of the Association
  for Computational Linguistics}, pages 7506--7516, Online. Association for
  Computational Linguistics.

\bibitem[{Jaradeh et~al.(2019)Jaradeh, Oelen, Farfar, Prinz, D'Souza,
  Kismihók, Stocker, and Auer}]{jaradeh2019orkg}
Mohamad~Yaser Jaradeh, Allard Oelen, Kheir~Eddine Farfar, Manuel Prinz,
  Jennifer D'Souza, Gábor Kismihók, Markus Stocker, and Sören Auer. 2019.
\newblock \href {https://doi.org/10.1145/3360901.3364435} {{Open Research
  Knowledge Graph}: Next generation infrastructure for semantic scholarly
  knowledge}.
\newblock In \emph{Proceedings of the 10th International Conference on
  Knowledge Capture}, pages 243--246. ACM.
\newblock \href{https://arxiv.org/abs/1901.10816}{Open-access manuscript}.

\bibitem[{Kim et~al.(2025)Kim, Suk, Cho, Longpre, Kim, Yoon, Son, Cho,
  Shafayat, Baek, Park, Hwang, Jo, Cho, Shin, Lee, Oh, Lee, Ho, Joo, Ko, Lee,
  Chae, Shin, Jang, Ye, Lin, Welleck, Neubig, Lee, Lee, and
  Seo}]{kim2025biggen}
Seungone Kim, Juyoung Suk, Ji~Yong Cho, Shayne Longpre, Chaeeun Kim, Dongkeun
  Yoon, Guijin Son, Yejin Cho, Sheikh Shafayat, Jinheon Baek, Sue~Hyun Park,
  Hyeonbin Hwang, Jinkyung Jo, Hyowon Cho, Haebin Shin, Seongyun Lee, Hanseok
  Oh, Noah Lee, Namgyu Ho, and 13 others. 2025.
\newblock \href {https://doi.org/10.18653/v1/2025.naacl-long.303} {The
  {B}i{GG}en bench: A principled benchmark for fine-grained evaluation of
  language models with language models}.
\newblock In \emph{Proceedings of the 2025 Conference of the Nations of the
  Americas Chapter of the Association for Computational Linguistics: Human
  Language Technologies (Volume 1: Long Papers)}, pages 5877--5919,
  Albuquerque, New Mexico. Association for Computational Linguistics.

\bibitem[{Koehn(2004)}]{koehn2004bootstrap}
Philipp Koehn. 2004.
\newblock \href {https://aclanthology.org/W04-3250/} {Statistical significance
  tests for machine translation evaluation}.
\newblock In \emph{Proceedings of the 2004 Conference on Empirical Methods in
  Natural Language Processing}, pages 388--395, Barcelona, Spain. Association
  for Computational Linguistics.

\bibitem[{Lebo et~al.(2013)Lebo, Sahoo, and McGuinness}]{lebo2013provo}
Timothy Lebo, Satya Sahoo, and Deborah McGuinness. 2013.
\newblock \href {https://www.w3.org/TR/prov-o/} {{PROV-O}: The {PROV}
  ontology}.
\newblock {W3C} recommendation, World Wide Web Consortium.

\bibitem[{Lewis et~al.(2020)Lewis, Perez, Piktus, Petroni, Karpukhin, Goyal,
  Küttler, Lewis, Yih, Rocktäschel, Riedel, and Kiela}]{lewis2020rag}
Patrick Lewis, Ethan Perez, Aleksandra Piktus, Fabio Petroni, Vladimir
  Karpukhin, Naman Goyal, Heinrich Küttler, Mike Lewis, Wen-tau Yih, Tim
  Rocktäschel, Sebastian Riedel, and Douwe Kiela. 2020.
\newblock \href
  {https://proceedings.neurips.cc/paper/2020/hash/6b493230205f780e1bc26945df7481e5-Abstract.html}
  {Retrieval-augmented generation for knowledge-intensive {NLP} tasks}.
\newblock In \emph{Advances in Neural Information Processing Systems},
  volume~33, pages 9459--9474. Curran Associates, Inc.

\bibitem[{Liakata et~al.(2012)Liakata, Saha, Dobnik, Batchelor, and
  Rebholz-Schuhmann}]{liakata2012coresc}
Maria Liakata, Shyamasree Saha, Simon Dobnik, Colin Batchelor, and Dietrich
  Rebholz-Schuhmann. 2012.
\newblock \href {https://doi.org/10.1093/bioinformatics/bts071} {Automatic
  recognition of conceptualization zones in scientific articles and two life
  science applications}.
\newblock \emph{Bioinformatics}, 28(7):991--1000.

\bibitem[{Liu et~al.(2024)Liu, Lin, Hewitt, Paranjape, Bevilacqua, Petroni, and
  Liang}]{liu2024lost}
Nelson~F. Liu, Kevin Lin, John Hewitt, Ashwin Paranjape, Michele Bevilacqua,
  Fabio Petroni, and Percy Liang. 2024.
\newblock \href {https://doi.org/10.1162/tacl_a_00638} {Lost in the middle: How
  language models use long contexts}.
\newblock \emph{Transactions of the Association for Computational Linguistics},
  12:157--173.

\bibitem[{Liu et~al.(2023)Liu, Iter, Xu, Wang, Xu, and Zhu}]{liu2023geval}
Yang Liu, Dan Iter, Yichong Xu, Shuohang Wang, Ruochen Xu, and Chenguang Zhu.
  2023.
\newblock \href {https://doi.org/10.18653/v1/2023.emnlp-main.153} {{G}-eval:
  {NLG} evaluation using gpt-4 with better human alignment}.
\newblock In \emph{Proceedings of the 2023 Conference on Empirical Methods in
  Natural Language Processing}, pages 2511--2522, Singapore. Association for
  Computational Linguistics.

\bibitem[{Lo et~al.(2020)Lo, Wang, Neumann, Kinney, and Weld}]{lo2020s2orc}
Kyle Lo, Lucy~Lu Wang, Mark Neumann, Rodney Kinney, and Daniel Weld. 2020.
\newblock \href {https://doi.org/10.18653/v1/2020.acl-main.447} {{S}2{ORC}: The
  semantic scholar open research corpus}.
\newblock In \emph{Proceedings of the 58th Annual Meeting of the Association
  for Computational Linguistics}, pages 4969--4983, Online. Association for
  Computational Linguistics.

\bibitem[{Lopez(2009)}]{lopez2009grobid}
Patrice Lopez. 2009.
\newblock \href {https://doi.org/10.1007/978-3-642-04346-8_62} {{GROBID}:
  Combining automatic bibliographic data recognition and term extraction for
  scholarship publications}.
\newblock In \emph{Research and Advanced Technology for Digital Libraries},
  pages 473--474. Springer.

\bibitem[{Luan et~al.(2018)Luan, He, Ostendorf, and
  Hajishirzi}]{luan2018scierc}
Yi~Luan, Luheng He, Mari Ostendorf, and Hannaneh Hajishirzi. 2018.
\newblock \href {https://doi.org/10.18653/v1/D18-1360} {Multi-task
  identification of entities, relations, and coreference for scientific
  knowledge graph construction}.
\newblock In \emph{Proceedings of the 2018 Conference on Empirical Methods in
  Natural Language Processing}, pages 3219--3232, Brussels, Belgium.
  Association for Computational Linguistics.

\bibitem[{Lála et~al.(2023)Lála, O'Donoghue, Shtedritski, Cox, Rodriques, and
  White}]{lala2023paperqa}
Jakub Lála, Odhran O'Donoghue, Aleksandar Shtedritski, Sam Cox, Samuel~G.
  Rodriques, and Andrew~D. White. 2023.
\newblock \href {https://arxiv.org/abs/2312.07559} {{PaperQA}:
  Retrieval-augmented generative agent for scientific research}.
\newblock \emph{Preprint}, arXiv:2312.07559.

\bibitem[{{Meta}(2025)}]{meta2025llama4}
{Meta}. 2025.
\newblock \href
  {https://huggingface.co/meta-llama/Llama-4-Scout-17B-16E-Instruct} {{Llama 4
  Scout} model card}.

\bibitem[{{National Information Standards Organization}(2024)}]{niso2024jats}
{National Information Standards Organization}. 2024.
\newblock \href {https://jats.nlm.nih.gov/publishing/1.4/} {{ANSI/NISO
  Z39.96-2024: Journal Article Tag Suite (JATS), Version 1.4}}.

\bibitem[{Oelen et~al.(2024)Oelen, Jaradeh, and Auer}]{oelen2024orkgask}
Allard Oelen, Mohamad~Yaser Jaradeh, and Sören Auer. 2024.
\newblock \href {https://arxiv.org/abs/2412.04977} {{ORKG ASK}: a
  neuro-symbolic scholarly search and exploration system}.
\newblock \emph{Preprint}, arXiv:2412.04977.

\bibitem[{{Open Research Knowledge Graph}(n.d.)}]{orkgtemplates}
{Open Research Knowledge Graph}. n.d.
\newblock \href
  {https://academy.orkg.org/tutorials/template-creation-tutorial.html}
  {Creating templates in the {ORKG}}.
\newblock Accessed 2026-10-11.

\bibitem[{Ren et~al.(2025)Ren, Rammohan, Scheck, Nguyen, Li, and
  Schultz}]{ren2025unlearning}
Zhao Ren, Rathi~Adarshi Rammohan, Kevin Scheck, Thanh~Tam Nguyen, Sheng Li, and
  Tanja Schultz. 2025.
\newblock \href {https://arxiv.org/abs/2510.04251} {Retain-free machine
  unlearning for speech emotion recognition}.
\newblock \emph{Preprint}, arXiv:2510.04251.

\bibitem[{{Research Object Community}(2026)}]{rocrate2026spec}
{Research Object Community}. 2026.
\newblock \href {https://doi.org/10.5281/zenodo.20720080} {{RO-Crate} metadata
  specification 1.3}.
\newblock Recommendation, 22 June 2026.

\bibitem[{Skarlinski et~al.(2024)Skarlinski, Cox, Laurent, Braza, Hinks,
  Hammerling, Ponnapati, Rodriques, and White}]{skarlinski2024paperqa2}
Michael~D. Skarlinski, Sam Cox, Jon~M. Laurent, James~D. Braza, Michaela Hinks,
  Michael~J. Hammerling, Manvitha Ponnapati, Samuel~G. Rodriques, and Andrew~D.
  White. 2024.
\newblock \href {https://arxiv.org/abs/2409.13740} {Language agents achieve
  superhuman synthesis of scientific knowledge}.
\newblock \emph{Preprint}, arXiv:2409.13740.

\bibitem[{Soiland-Reyes et~al.(2022)Soiland-Reyes, Sefton, Crosas, Castro,
  Coppens, Fernández, Garijo, Grüning, La~Rosa, Leo, Ó~Carragáin, Portier,
  Trisovic, {RO-Crate Community}, Groth, and Goble}]{soilandreyes2022rocrate}
Stian Soiland-Reyes, Peter Sefton, Mercè Crosas, Leyla~Jael Castro, Frederik
  Coppens, José~M. Fernández, Daniel Garijo, Björn Grüning, Marco La~Rosa,
  Simone Leo, Eoghan Ó~Carragáin, Marc Portier, Ana Trisovic, {RO-Crate
  Community}, Paul Groth, and Carole Goble. 2022.
\newblock \href {https://doi.org/10.3233/DS-210053} {Packaging research
  artefacts with {RO-Crate}}.
\newblock \emph{Data Science}, 5(2):97--138.

\bibitem[{Song et~al.(2025)Song, Shen, Hou, Wang, Wang, Tang, and
  Lv}]{song2025fedaghn}
Jiarui Song, Yunheng Shen, Chengbin Hou, Pengyu Wang, Jinbao Wang, Ke~Tang, and
  Hairong Lv. 2025.
\newblock \href {https://arxiv.org/abs/2501.16379v2} {Fedaghn: Personalized
  federated learning with attentive graph hypernetworks}.
\newblock \emph{Preprint}, arXiv:2501.16379v2.

\bibitem[{Teufel et~al.(2009)Teufel, Siddharthan, and Batchelor}]{teufel2009az}
Simone Teufel, Advaith Siddharthan, and Colin Batchelor. 2009.
\newblock \href {https://aclanthology.org/D09-1155/} {Towards
  discipline-independent argumentative zoning: Evidence from chemistry and
  computational linguistics}.
\newblock In \emph{Proceedings of the 2009 Conference on Empirical Methods in
  Natural Language Processing}, pages 1493--1502, Singapore. Association for
  Computational Linguistics.

\bibitem[{Wadden et~al.(2019)Wadden, Wennberg, Luan, and
  Hajishirzi}]{wadden2019dygiepp}
David Wadden, Ulme Wennberg, Yi~Luan, and Hannaneh Hajishirzi. 2019.
\newblock \href {https://doi.org/10.18653/v1/D19-1585} {Entity, relation, and
  event extraction with contextualized span representations}.
\newblock In \emph{Proceedings of the 2019 Conference on Empirical Methods in
  Natural Language Processing and the 9th International Joint Conference on
  Natural Language Processing (EMNLP-IJCNLP)}, pages 5784--5789, Hong Kong,
  China. Association for Computational Linguistics.

\bibitem[{Wang et~al.(2024)Wang, Xu, Zhao, Ouyang, Wu, Zhao, Xu, Liu, Qu,
  Shang, Zhang, Wei, Sui, Li, Shi, Qiao, Lin, and He}]{wang2024mineru}
Bin Wang, Chao Xu, Xiaomeng Zhao, Linke Ouyang, Fan Wu, Zhiyuan Zhao, Rui Xu,
  Kaiwen Liu, Yuan Qu, Fukai Shang, Bo~Zhang, Liqun Wei, Zhihao Sui, Wei Li,
  Botian Shi, Yu~Qiao, Dahua Lin, and Conghui He. 2024.
\newblock \href {https://arxiv.org/abs/2409.18839} {{MinerU}: An open-source
  solution for precise document content extraction}.
\newblock \emph{Preprint}, arXiv:2409.18839.

\bibitem[{Wang et~al.(2026)Wang, Li, Fan, Chen, Chen, Zhou, and
  Zhou}]{wang2026scicore}
Chenguang Wang, Ming Li, Chengrui Fan, Jianpeng Chen, Han Chen, Tianyi Zhou,
  and Dawei Zhou. 2026.
\newblock \href {https://arxiv.org/abs/2609.39027} {A missing piece for
  trustworthy {AI} reviewers: From benchmarking rhetorical robustness to
  {SciCore} review}.
\newblock \emph{Preprint}, arXiv:2609.39027.

\bibitem[{Wu et~al.(2026)Wu, Zhang, Li, Xu, Duan, Liu, Pan, Zhu, Zhou, Wei, Li,
  Chen, He, and Tan}]{wu2026internatlas}
Yujun Wu, Dongxu Zhang, Xinchen Li, Jinhang Xu, Yiling Duan, Yumou Liu, Jiabao
  Pan, Qiyuan Zhu, Xuanhe Zhou, Jingxuan Wei, Siyuan Li, Jintao Chen, Conghui
  He, and Cheng Tan. 2026.
\newblock \href {https://doi.org/10.48550/arXiv.2604.28158} {{Intern-Atlas}: A
  methodological evolution graph as research infrastructure for ai scientists}.
\newblock \emph{Preprint}, arXiv:2604.28158v2.

\bibitem[{Zhang et~al.(2024)Zhang, Chen, Pan, Caragea, Latecki, and
  Dragut}]{zhang2024scier}
Qi~Zhang, Zhijia Chen, Huitong Pan, Cornelia Caragea, Longin~Jan Latecki, and
  Eduard Dragut. 2024.
\newblock \href {https://doi.org/10.18653/v1/2024.emnlp-main.726} {{S}ci{ER}:
  An entity and relation extraction dataset for datasets, methods, and tasks in
  scientific documents}.
\newblock In \emph{Proceedings of the 2024 Conference on Empirical Methods in
  Natural Language Processing}, pages 13083--13100, Miami, Florida, USA.
  Association for Computational Linguistics.

\bibitem[{Zheng et~al.(2023)Zheng, Chiang, Sheng, Zhuang, Wu, Zhuang, Lin, Li,
  Li, Xing, Zhang, Gonzalez, and Stoica}]{zheng2023judge}
Lianmin Zheng, Wei-Lin Chiang, Ying Sheng, Siyuan Zhuang, Zhanghao Wu, Yonghao
  Zhuang, Zi~Lin, Zhuohan Li, Dacheng Li, Eric Xing, Hao Zhang, Joseph
  Gonzalez, and Ion Stoica. 2023.
\newblock \href {https://doi.org/10.52202/075280-2020} {Judging
  {LLM}-as-a-judge with mt-bench and chatbot arena}.
\newblock In \emph{Advances in Neural Information Processing Systems},
  volume~36, pages 46595--46623. Curran Associates, Inc.

\bibitem[{Zhu et~al.(2025)Zhu, Xie, Liu, Li, and Hu}]{zhu2025kg2rag}
Xiangrong Zhu, Yuexiang Xie, Yi~Liu, Yaliang Li, and Wei Hu. 2025.
\newblock \href {https://doi.org/10.18653/v1/2025.naacl-long.449} {Knowledge
  graph-guided retrieval augmented generation}.
\newblock In \emph{Proceedings of the 2025 Conference of the Nations of the
  Americas Chapter of the Association for Computational Linguistics: Human
  Language Technologies (Volume 1: Long Papers)}, pages 8912--8924,
  Albuquerque, New Mexico. Association for Computational Linguistics.

\end{thebibliography}
\endgroup

\clearpage
\appendix
\setcounter{dbltopnumber}{3}
\renewcommand{\dbltopfraction}{0.85}
\section{Resource and Construction}

\subsection{Resource Scope and Sampling}
\label{app:resource-statistics}

The corpus contains 1,059,112 SPRA paper representations spanning 1965--2026, produced through the construction procedure in Appendix~\ref{app:construction-spec} and exposed through Sciverse. The source-aligned evaluation sample contains 9,849 eligible papers published in 2025, with 1,407 per field across seven AI subfields (Table~\ref{tab:subfield-counts}). The cohort follows the primary Reader's documented August 2024 knowledge cutoff \citep{meta2025llama4}. The sampled SPRA instances retain their corpus research objects, relations, and source bindings, linking evaluation to the resource's production workflow. The resulting assessments characterize the eligible 2025 cohort across these seven fields. RQ3 evaluates 1,001 task variants across seven AI domains, as detailed in Appendix~\ref{app:additional-results}.

Corpus statistics describe the collection used in this study, measured in September 2026. The supplementary records specify the counting populations and preserve the aggregate results and evaluation data. The Sciverse Paper Schema API supports continued paper- and object-level access; Appendix~\ref{app:reproducibility} describes the accompanying evaluation materials for offline inspection and numerical reanalysis.

\begin{table*}[!b]
\centering\small

\begin{tabular}{@{}L{0.78\textwidth}r@{}}
\toprule
AI subfield & Eligible \\
\midrule
Language, speech, and information access & 1,407 \\
Vision and multimodal perception & 1,407 \\
Learning, optimization, and general ML & 1,407 \\
Data, knowledge, and computational systems & 1,407 \\
Reasoning, agents, RL, and embodied AI & 1,407 \\
Trustworthy and human-centered AI & 1,407 \\
AI for science and domain applications & 1,407 \\
\midrule
Total & 9,849 \\
\bottomrule
\end{tabular}
\caption{Eligible evaluation papers across seven AI sampling strata.}
\label{tab:subfield-counts}
\end{table*}

The corpus contains 19,949,289 core units (18.84/paper) and 24,073,381 internal relations (22.73/paper). Core units comprise Problems, Contributions, Components, ExperimentSetups, Measures, and Findings; relation counts cover the eight profile predicates and exclude citations. Machine-readable statistics report per-type counts and the year distribution.

The sample contains 214,239 core units: 9,853 Problems, 11,395 Contributions, 36,522 Components, 37,981 ExperimentSetups, 44,355 Measures, and 74,133 Findings (mean 21.75/paper; median 21; range 4--213). Separate counts are 9,849 Documents, 177,003 References, and 8,646 Resources. Its 409,082 locator references count repeated addresses separately.

Complete relation retrieval covers 4,350 papers and 109,591 relations (mean 25.19/paper; median 24; range 4--295). These complete groups define the structural-check population, rather than a second balanced sample.

\subsection{Sources, Acquisition, and Resource Access}
\label{app:resource-access}

\paragraph{Sources and acquisition.} The evaluation roster includes AAAI, ACL, EMNLP, ICLR, ICML, NeurIPS, CVPR, ICCV, robotics proceedings, and arXiv. Venue labels identify publication outlets; recorded arXiv, ACL Anthology, AAAI, and DOI-linked URLs identify accessed versions. Sciverse supplies metadata, parsed sources, and SPRA instances; MinerU produces section-aware Markdown. The packaged statistics and source-linked examples are grouped in \path{data/data/supplementary/paper_results.json} and \path{data/data/supplementary/frozen_subset.json}; the Sciverse Paper Schema API provides paper- and object-level corpus access.

\paragraph{Service and data access.} The accompanying data archive contains the complete 9,849-paper aligned RQ1 cohort with all three input representations: complete text, retrieved passages, and SPRA. Together with the profile, attributed examples, corpus statistics, frozen study outputs, and analysis scripts, this supports offline inspection and numerical replay without service credentials (Appendix~\ref{app:artifact}). Sciverse exposes the million-record corpus through the token-authenticated \href{https://sciverse.opendatalab.com/docs/sciverse/api/paper-schema}{Sciverse Paper Schema API} for paper discovery and access to typed units, relations, citations, and source context. The API supports discovery by paper metadata or research objects, followed by goal-specific material selection for overview, survey, benchmark, method, and reproduction tasks. Agents can retrieve typed records, follow relations and resolved citation links, and reopen source context. Material responses report available and returned counts, including truncation, so callers can request additional records when needed. The Paper Schema API documentation specifies request and response formats; API Tokens are created through the \href{https://sciverse.opendatalab.com/}{Sciverse} account console. The \href{https://sciverse.opendatalab.com/docs/sciverse/skills/paper-schema}{Paper Schema Agent Skill documentation} describes the corresponding tools and installation. Access is subject to account permissions and quotas.

\paragraph{Licenses.} Project software uses Apache-2.0; project-authored benchmark records, labels, prompts, and documentation use CC BY 4.0. Third-party articles and service payloads retain their source terms; redistribution follows the terms of each source. The seven inspection instances retain CC-BY attribution. Evaluation metadata records 8,340 positive, 569 negative, and 940 unspecified open-access flags, separately from redistribution permissions in the source-version access records.

\subsection{Construction Procedure}
\label{app:construction-spec}

MinerU supplies Markdown with paragraph markers and verbatim tables. Five core calls share a paper-owned registry: an initial extraction stage assigns unit identifiers and source markers; a structural pass predicts \texttt{part\_of}, \texttt{co\_contribution}, \texttt{compares\_to}, and \texttt{evaluates}; three role-specific passes enrich Problem, Method, and Evidence units and predict relations such as \texttt{motivates}, \texttt{about}, and \texttt{supports}. The Evidence pass creates Findings and binds their measures and conditions. A \texttt{supports} edge connects a Finding to associated research objects, giving agents direct access to evidence-bearing content and, where recorded, its source locations.

Separate metadata, citation-role, and reference-metadata calls form the bibliographic and external-dependency layers. Deterministic assembly reconciles identifiers, merges records, derives \texttt{resolves}, and validates fields, endpoints, ownership, and unit coverage. Prompts constrain admissible types, identifier reuse, and evidence; citation keys join external records without merging them into internal claims.

The \href{https://github.com/gavinwxy/ai-paper-structured-extraction}{SPRA extraction implementation} uses a shared paper-level registry and role-specific extraction stages. Its model configuration and request settings are summarized in Appendix~\ref{app:model-call-snapshot} and the supplementary registry.

\paragraph{Validation.} Required extraction stages must produce schema-valid records; unresolved validation issues exclude a paper from task use. Assembly normalizes identifiers and source markers and removes duplicate, dangling, self-loop, unknown-predicate, and invalid-signature edges. Metadata or citation-extraction errors are retained with warnings when core extraction remains usable.

\paragraph{Endpoint contract.} For \texttt{compares\_to}, both source and target types belong to \{Contribution, Component, ExperimentSetup, Measure\}. For \texttt{motivates}, the source is a Problem and the target is a Contribution, Component, or ExperimentSetup. Every internal endpoint must belong to the same paper's unit registry. Table~\ref{tab:relations} lists the eight internal relation types and their admissible endpoint signatures.

Source markers are normalized against parsed blocks and headings. Tables are sliced verbatim. An optional numeric-presence checker flags Measure values absent from captured table cells, records unchecked cases, and leaves units and scores unchanged. The construction rules require preservation of values, units, comparison bases, and conditions; numeric presence alone does not verify these field associations.

\subsection{Structural Validation}
\label{app:construction-quality}
\label{app:construction-audit}

These checks quantify two prerequisites for inspecting represented claims: source traceability and relation endpoint connectivity. Among 214,239 sampled research units, 203,628 (95.0\%) have a nonempty provenance field. In the complete relation records for 4,350 papers, 106,439 of 109,591 relations (97.1\%) join to both endpoints; 102,024 of these joined relations have provenance. These measures characterize the availability of source pointers and connected relation endpoints for agent access. The \texttt{supports} relation provides evidence links between research objects and Findings; agents can follow these links to retrieve associated content and inspect the source evidence when available. RQ1 complements these checks by evaluating the scientific use of extracted content in source-conditioned answers.

A sampled object assessment provides complementary record-level evidence; the supplementary materials document its protocol and records.

\subsection{Contribution-Level Semantic Alignment}
\label{app:ncg-semantic}

A paper representation should preserve the scientific contributions expressed in the source article. We assess this property against the public annotations of NCG/SemEval 2021 Task 11 \citep{dsouza2021semeval}, comparing SPRA objects with annotated contribution sentences, concepts, information-unit families, and contribution facts.

The projection implementation reads precomputed extraction records and maps their types and content to NCG information-unit families: Problem objects supply research problems, ExperimentSetup objects supply experimental settings, and Finding and Measure objects supply results. For each gold contribution sentence, it selects the extracted string with the highest token F1 as a candidate alignment. GPT-4.1 receives the gold annotations and projected content, then judges semantic correspondence rather than lexical overlap. Thus, candidate matching organizes the evidence for a gold-conditioned coverage assessment. The rubric accepts paraphrase and canonicalization, and penalizes unsupported additions, incorrect attribution, and omitted contribution facts. Five dimensions receive scores from 0 to 5, alongside a separately judged Overall score from 0 to 100. These paper-level judgments measure the preservation of annotated contribution structure.

\begin{table}[t]
\centering\small

\begin{tabular}{@{}lr@{}}
\toprule
Dimension & Mean \\
\midrule
Contribution-sentence coverage & 3.967 \\
Key-concept coverage & 4.031 \\
Information-unit alignment & 4.436 \\
Contribution-fact coverage & 3.890 \\
Evidence traceability & 4.998 \\
\midrule
Overall (0--100) & 84.49 \\
\bottomrule
\end{tabular}
\caption{NCG contribution-level semantic assessment. Dimension scores use 0--5; Overall is a separately returned model score on 0--100.}
\label{tab:ncg-semantic}
\end{table}

SPRA achieves a mean Overall score of 84.49 and a median of 85. Information-unit alignment scores higher (4.436/5) than contribution-fact coverage (3.890/5; Table~\ref{tab:ncg-semantic}). This pattern suggests that SPRA captures the organization of scientific contributions more consistently than their fine-grained factual detail. The assessment complements RQ1 by examining correspondence with annotated scientific content before its use in downstream answers.

The supplement provides the projection and judge implementations, rubric, aggregate scores, domain summaries, and paper-level call metadata. Complete run-specific judge inputs and individual judgments are unavailable; the retained code documents the procedure, while the accompanying NCG documentation identifies the additional records needed to reproduce the reported aggregate scores.

\subsection{Structural Comparison of Scholarly Representations}
\label{app:representation-comparison}

Table~\ref{tab:representation-features} compares documented capabilities of scholarly representations and infrastructure. JATS addresses article markup \citep{niso2024jats}; PROV-O defines the provenance ontology \citep{lebo2013provo}, while PROV-CONSTRAINTS specifies validity constraints \citep{w3c2013provconstraints}; RO-Crate addresses research-object packaging \citep{rocrate2026spec}. SciREX and NLPContributionGraph (NCG) address scientific extraction, while ORKG provides a maintained scholarly graph \citep{jain2020scirex,dsouza2021semeval,jaradeh2019orkg,orkgtemplates,oelen2024orkgask}. The comparison concerns their published scope, rather than measured extraction accuracy.

\begin{table*}[t]
\centering\footnotesize
\renewcommand{\arraystretch}{1.15}

\begin{tabular*}{\textwidth}{@{\extracolsep{\fill}}lccccccc@{}}
\toprule
Capability & \shortstack{JATS\\1.4} & ORKG & PROV-O & \shortstack{RO-Crate\\v1.3} & SciREX & NCG & \textbf{SPRA} \\
\midrule
Paper-associated research units & $\triangle$ & $\checkmark$ & $\triangle$ & $\triangle$ & $\checkmark$ & $\checkmark$ & $\checkmark$ \\
Typed scientific relations & $\triangle$ & $\checkmark$ & $\triangle$ & $\triangle$ & $\checkmark$ & $\checkmark$ & $\checkmark$ \\
Units linked to source content & $\triangle$ & $\triangle$ & $\triangle$ & $\triangle$ & $\checkmark$ & $\checkmark$ & $\checkmark$ \\
Persistent resource identifiers & $\triangle$ & $\checkmark$ & $\checkmark$ & $\checkmark$ & $\triangle$ & $\triangle$ & $\checkmark$ \\
Explicit schema/profile validation & $\checkmark$ & $\checkmark$ & $\triangle$ & $\checkmark$ & $\triangle$ & $\triangle$ & $\checkmark$ \\
Scientific lookup/traversal API & $\times$ & $\checkmark$ & $\times$ & $\times$ & $\times$ & $\triangle$ & $\checkmark$ \\
Matched downstream task evaluation & $\times$ & $\triangle$ & $\times$ & $\times$ & $\times$ & $\times$ & $\checkmark$ \\
Condition/qualification access & $\triangle$ & $\triangle$ & $\triangle$ & $\triangle$ & $\triangle$ & $\triangle$ & $\checkmark$ \\
Version/revision metadata & $\checkmark$ & $\checkmark$ & $\checkmark$ & $\checkmark$ & $\triangle$ & $\triangle$ & $\checkmark$ \\
\bottomrule
\end{tabular*}
\caption{Feature coverage of adjacent scholarly representations. $\checkmark$ denotes documented core support, $\triangle$ partial support or support requiring a profile, integration, or extension, and $\times$ no documented core support for the stated capability. NCG = NLPContributionGraph.}
\label{tab:representation-features}
\end{table*}

\paragraph{Schema coverage and granularity.} SciREX represents four entity types---Method, Task, Dataset, and Metric---with mentions, coreference, salience, and document-level $n$-ary result relations. NCG organizes contribution sentences, phrases, and triples under 12 information units and records a \texttt{from sentence} association. Both therefore already associate extracted content with a paper and its text. SPRA's six-role empirical-AI profile instead makes the research argument explicit: Problems motivate Contributions, Components and ExperimentSetups describe implementation and evaluation context, and Measures connect to Findings with conditions. Its eight internal predicates have admissible endpoint signatures (Table~\ref{tab:relations}). Type counts describe granularity, not a coverage or quality ranking.

\paragraph{From scientific tasks to the profile.} Task-derived roles express a design mapping from scientific operations to representation requirements. Paper-understanding questions call for Problem and Contribution objects and their \texttt{motivates}/\texttt{resolves} links. Method comparison calls for Components and ExperimentSetups, membership, and comparison edges. Interpreting reported evidence calls for Measures and Findings, \texttt{evaluates}/\texttt{about}/\texttt{supports} links, and qualification attributes. Here, \texttt{supports} is an extracted candidate evidence association for navigation; semantic support is assessed by comparing a recorded source locator, when present, with the Finding. Verification requires source locators; reuse requires stable paper-scoped identifiers. This is an explicit task-to-profile design rationale, rather than an automated schema-induction procedure.

\paragraph{Persistence, source binding, and operations.} ORKG already provides persistent, paper-linked contributions, templates with ranges and cardinalities, comparisons, and APIs. Paper ownership is a shared design property of ORKG and SPRA. Source binding also differs in granularity: SciREX retains mention spans, NCG links triples to contribution sentences, and SPRA records addresses for passages, tables, and figures. ORKG can represent additional evidence and qualifiers through properties and templates. SPRA's contribution is the integration of a paper-level research-argument profile, constrained internal relations, source-opening operations, and a common object registry into a corpus-scale extraction and access workflow. Typed lookup, traversal, and filtering operate on these same objects; RQ1--RQ3 assess their scientific use under matched tasks. Matched downstream evaluation in the table refers to representation-access studies, distinct from extraction benchmarks. Property alignment connects this task-oriented representation to ORKG's broader aggregation infrastructure.

\subsection{ORKG Graph Sample and Interoperability Scope}
\label{app:orkg-export}

We examine interoperability using a sample of 200 papers from ORKG's Artificial Intelligence field and subfields. We select 200 papers from a 2,918-paper frame by deterministic ordering before inspecting graph properties. The sample includes 313 contribution roots and 6,578 statement occurrences, with contribution subgraphs traversed to three hops.

The mapping analysis examines how ORKG research-role classes and predicate labels correspond to the SPRA profile while preserving scientific identities, reference-paper links, and relation directions. The graph sample, mapping rules, and protocol are grouped in the ORKG section of \path{data/data/supplementary/paper_results.json}. These materials characterize the interoperability interface between the two representations.

\begin{table}[t]
\centering\footnotesize

\begin{tabular}{@{}L{0.31\columnwidth}L{0.61\columnwidth}@{}}
\toprule
Component & Description \\
\midrule
Sample frame & 2,918 listed papers; 200 selected deterministically \\
Graph contexts & 313 contribution roots; three-hop traversal \\
Scientific statements & 6,578 statement occurrences \\
Mapping rules & Explicit research-role classes and predicate labels \\
Source identity & Scientific resource identifiers and paper-reference links \\
\bottomrule
\end{tabular}
\caption{ORKG graph sample and mapping specification used to examine interoperability.}
\label{tab:orkg-export-workflow}
\end{table}

\section{Evaluation Details}
\label{app:additional-results}

\subsection{Protocol and Analysis Population}

\paragraph{Complete paired analysis sets.}
Table~\ref{tab:complete-case-population} gives the denominators behind the reported comparisons. RQ1 excludes 411 items without complete three-condition results: 383 stop at Reader generation, 21 at reference generation, and seven at primary judging. G-U3 is the Problem + Contribution subset of G-U1; it is not an additional task population.

\begin{table*}[t]
\centering\small

\begin{tabular}{@{}lrrr@{}}
\toprule
Study or outcome & Analysis frame & Complete pairs & Incomplete \\
\midrule
RQ1, all question families & 9,849 & 9,438 & 411 \\
RQ1 G-U1, semantic QA & 5,628 & 5,510 & 118 \\
RQ1 G-U2, structured operations & 4,221 & 3,928 & 293 \\
RQ1 G-U3, Problem + Contribution & 1,407 & 1,378 & 29 \\
RQ2, directed & 2,461 & 2,431 & 30 \\
RQ2, Top-5 & 961 & 639 & 322 \\
RQ3, proposal generation & 1,001 & 1,001 & 0 \\
\bottomrule
\end{tabular}
\caption{Analysis populations and complete paired observations. The frame is source-aligned for RQ1 and input-eligible for RQ2; RQ3 includes all planned variants. G-U3 is included within G-U1.}
\label{tab:complete-case-population}
\end{table*}

We compare representations using paired tasks under a common primary Reader and evaluation protocol. Evidence inputs are fixed before inference, and the model configurations and study records are provided in Appendix~\ref{app:model-call-snapshot}.

All RQ1--RQ3 and citation-expansion quality scores use the 0--5 scale. Judge dimensions recorded on the 0--4 rubric are reported as $s=5r/4$. Paired differences and their intervals are expressed in score points. Rates and input+output tokens use their stated units.

Llama 4 Scout is the primary Reader for RQ1--RQ3; two long-context RQ2 Full Compact tasks use the documented GPT fallback. DeepSeek-v4-pro generates shared RQ1/RQ2 answer references from Full Compact; MiniMax-M3, blind to condition, scores Reader responses against them. The RQ1 evaluation frame contains 9,849 items across seven AI subfields. RQ2 evaluates 3,070 complete paired tasks; RQ3 evaluates 1,001 variants. Table~\ref{tab:analysis-population} lists the primary comparisons.

\begin{table}[t]
\centering
\footnotesize

\begin{tabular}{@{}L{0.13\columnwidth}L{0.23\columnwidth}L{0.52\columnwidth}@{}}
\toprule
Study & Evaluation scope & Primary comparison \\
\midrule
RQ1 & 9,849 & Schema-RAG vs. Text-RAG (answer quality) \\
\addlinespace[1.5pt]
RQ2 & 3,070 complete pairs & Correctness: Schema-RAG vs. Text-RAG (directed);\newline Schema-RAG vs. Full Compact (Top-5) \\
\addlinespace[1.5pt]
RQ3 & 1,001 & Overall idea quality, three conditions \\
\bottomrule
\end{tabular}
\caption{Evaluation scope and primary comparisons.}
\label{tab:analysis-population}
\end{table}

SPRA names the representation; Schema-RAG names its access condition in this comparison. Task evidence is fixed before Reader inference. For RQ2, paper groups, source revisions, questions, and the per-paper, per-facet selection slots are shared across conditions; representation-specific input packs are frozen before inference, and retrieval is independent of generated answers. Both rankers use question-derived queries and BM25F-like field weighting to select one primary evidence item per paper for each of four scientific facets. Text-RAG ranks source passages; Schema-RAG ranks SPRA objects produced through schema-constrained extraction and deterministic assembly, together with typed internal relations. The shared slots match selection opportunities while preserving each representation's distinct evidence form and resulting input length.

\subsection{Controlled Representation Comparison}

\paragraph{Full Compact input boundary.}\label{app:full-compact-input}
Full Compact exposes the complete available parsed article text in source order. This includes table contents, formulas, and figure captions where retained in the textual extraction. Figure pixels and page images are not supplied as visual inputs; image links in the text are not dereferenced. Thus, completeness refers to the parsed textual representation, not to every visual element of the original PDF. For RQ1, this text supplies the source for shared reference answers. Text-RAG exposes selected passages; Schema-RAG exposes SPRA objects and typed internal relations. Llama 4 Scout answers from each assigned condition, and MiniMax-M3 scores those responses against the Full Compact-generated references. Questions, paper identities, response contracts, and scoring are fixed.

\paragraph{Full-text context handling.} RQ2 Full Compact concatenates complete section-aware article views in shared paper order. Groups that fail the input gate are excluded jointly rather than truncating the full-text condition into eligibility. The 32 input-gated groups are outside the 3,422-task input-eligible population. RQ3 evidence digestion processes at most one Full Compact paper per call and at most two papers per RAG call, with a 65,000-token batch gate. These limits govern admissible inputs; token accounting includes the complete requests.

\paragraph{RQ1 metric construction.} Native judge dimensions are scored from 0 to 4 and displayed as $s=5r/4$. G-U1 averages correctness, answer coverage, and qualification over semantic questions; G-U2 uses structured correctness on the three operation families; G-U3 applies the G-U1 scoring rule to Problem + Contribution questions. These are rubric-based model judgments against shared references. They measure task performance, rather than probabilities that individual extracted records are correct. On the 0--5 reporting scale, the retained G-U1 SPRA--Full Compact contrast is $-0.078$ (95\% CI [$-0.110,-0.046$]). The references can contain source details absent from either selected representation, so the comparison assesses retained information together with Reader performance. Token totals are reported within their respective studies.

\subsection{RQ2 Token Accounting}

\label{app:rq2-token-accounting}

Token accounting covers the same 3,070 scored tasks in all three conditions: 2,431 directed two-paper tasks and 639 Top-5 tasks. For each task and condition, we sum one retained Reader input and its corresponding output. Inputs include evidence, instructions, task metadata, output-format instructions, and chat formatting. Reference generation, judging, failed calls, and retries are outside this retained-call measure.

The shared offline BERT WordPiece tokenizer (\texttt{bert-base-uncased}; frozen accounting identifier \texttt{offline\_bert\_wordpiece\_}\allowbreak\texttt{conservative\_gate\_v1}) measures both input and output length. These standardized counts describe model-visible text volume; provider billing uses its own tokenization and input/output rates. In Full Compact / Text-RAG / Schema-RAG order, input totals are 199,460,508 / 10,678,861 / 6,626,310 and output totals are 3,801,534 / 2,617,712 / 2,885,511. Their sums are 203,262,042 / 13,296,573 / 9,511,821 tokens, reported as 203.26 / 13.30 / 9.51 million.

Reductions use the unrounded aggregate totals, $100(1-T_{\mathrm{SPRA}}/T_{\mathrm{baseline}})$: 28.5\% relative to Text-RAG and 95.3\% relative to Full Compact. Table~\ref{tab:rq2-token-distribution} separates the two task types and reports per-task input+output distributions. Means divide each group's total by its task count; pooled means weight tasks equally. The supplement provides the individual counts and unrounded summaries.

\begin{table*}[t]
\centering\footnotesize

\begin{tabular*}{\textwidth}{@{\extracolsep{\fill}}llrrrr@{}}
\toprule
Tasks ($n$) & Condition & Mean & Median & Q1--Q3 & P90 \\
\midrule
All (3,070) & Full Compact & 66,209 & 56,093 & 39,863--88,487 & 120,199 \\
All (3,070) & Text-RAG & 4,331 & 3,550 & 3,204--4,157 & 7,591 \\
All (3,070) & Schema-RAG & 3,098 & 2,557 & 2,438--2,773 & 5,335 \\
\addlinespace[2pt]
Directed (2,431) & Full Compact & 54,032 & 48,610 & 36,786--64,701 & 85,691 \\
Directed (2,431) & Text-RAG & 3,472 & 3,413 & 3,125--3,704 & 3,973 \\
Directed (2,431) & Schema-RAG & 2,511 & 2,506 & 2,412--2,604 & 2,704 \\
\addlinespace[2pt]
Top-5 (639) & Full Compact & 112,534 & 114,423 & 100,636--127,501 & 136,319 \\
Top-5 (639) & Text-RAG & 7,600 & 7,535 & 7,155--7,968 & 8,463 \\
Top-5 (639) & Schema-RAG & 5,333 & 5,323 & 5,156--5,508 & 5,687 \\
\bottomrule
\end{tabular*}
\caption{RQ2 input+output tokens per retained Reader call. Each task contributes one call per condition. Q1--Q3 denotes the interquartile interval; P90 is the 90th percentile. Quantiles use linear interpolation; displayed values are rounded to the nearest token. Schema-RAG is SPRA-based access.}
\label{tab:rq2-token-distribution}
\end{table*}

\subsection{RQ3 Idea-Generation Quality and Token Use}

RQ3 evaluates 1,001 idea-generation variants across seven AI domains.

Full Compact supplies complete text, Text-RAG supplies retrieved passages, and Schema-RAG supplies SPRA research objects and internal relations. These conditions use the same paper selections and order.

Tables~\ref{tab:rq3-dimensions} and~\ref{tab:rq3-contrasts} report the question-weighted means and family-balanced paired tests. Overall is independently judged rather than averaged from dimensions. Intervals use a 50,000-draw benchmark-family cluster bootstrap; paired tests enumerate all $2^{14}$ family sign flips.

\begin{table}[t]
\centering\footnotesize

\begin{tabular}{@{}L{0.26\linewidth}L{0.43\linewidth}r@{}}
\toprule
Access condition & Input representation & Overall (0--5) \\
\midrule
Full Compact & Complete text & 2.965 \\
Text-RAG & Retrieved passages & 2.886 \\
Schema-RAG & SPRA research objects and relations & 2.898 \\
\bottomrule
\end{tabular}
\caption{RQ3 conditions and independently judged Overall means across 1,001 variants (0--5). Means weight variants equally; the paired contrasts in Table~\ref{tab:rq3-contrasts} weight benchmark families equally.}
\label{tab:rq3-dimensions}
\end{table}

\begin{table*}[t]
\centering\small

\begin{tabular*}{\textwidth}{@{\extracolsep{\fill}}lrrrr@{}}
\toprule
Contrast & Difference & 95\% cluster-bootstrap interval & Exact $p$ & Holm $p$ \\
\midrule
Schema-RAG--Full Compact & $-0.067$ & $[-0.102,-0.034]$ & 0.0020 & 0.0059 \\
Schema-RAG--Text-RAG & $+0.011$ & $[-0.029,+0.058]$ & 0.6792 & 0.6792 \\
Text-RAG--Full Compact & $-0.078$ & $[-0.129,-0.031]$ & 0.0096 & 0.0193 \\
\bottomrule
\end{tabular*}
\caption{RQ3 paired Overall contrasts across 1,001 variants, expressed in score points. Intervals use the 50,000-draw benchmark-family cluster bootstrap; exact $p$ values enumerate all 14-family sign flips. Each of the 14 benchmark families has equal weight in the paired estimate; Holm correction covers the three contrasts.}
\label{tab:rq3-contrasts}
\end{table*}

\paragraph{Input+output accounting.}
Table~\ref{tab:rq3-tokens} sums input and output tokens for digestion, generation, and binding. Shared selection, reference generation, auditing, and review are outside this task-generation measure.

\begin{table*}[t]
\centering\small

\begin{tabular*}{\textwidth}{@{\extracolsep{\fill}}lrr@{}}
\toprule
Workflow & Input+output (millions) & Reduction \\
\midrule
Full Compact & 155.92 & -- \\
Text-RAG & 11.87 & 92.4\% \\
Schema-RAG & 9.77 & 93.7\% \\
\bottomrule
\end{tabular*}
\caption{RQ3 input+output token accounting across the generation workflow. Reductions are relative to Full Compact.}
\label{tab:rq3-tokens}
\end{table*}

Storage, retrieval, and maintenance are supplied by Sciverse.

\paragraph{Evaluation scope.} The studies use shared benchmark data and evaluators to define inference clusters. Some tasks transfer benchmark task forms to public datasets rather than reproduce complete source experiments.

\subsection{Statistical Inference}

RQ1 uses 10,000 domain--family stratified paired-bootstrap replicates, preserving the paired-comparison principle \citep{koehn2004bootstrap}. RQ2 uses 10,000 multi-membership paper-cluster percentile replicates: unique papers are sampled multinomially, and each task receives the mean multiplicity of its member papers. Both preserve within-task condition pairing. RQ3 averages the three Overall reviews within each task--condition and equally weights cluster means, with each cluster defined by shared benchmark data and an evaluator. The statistical analysis uses 50,000 benchmark-cluster percentile-bootstrap draws, enumerates all paired cluster sign flips for the exact test, and applies Holm correction across the three Overall contrasts \citep{holm1979multiple}.

The paired analyses use tasks with results under all three conditions and the specified evaluators. Appendix~\ref{app:artifact} describes numerical replay and supporting materials.

\subsection{Citation-Guided Research Integration}
\label{app:relation-expansion}

Across 1,001 variants from 14 benchmark families and seven domains, both conditions read SPRA objects from the first two condition-neutral selected papers. The expanded condition additionally opens up to two resolved outgoing one-hop citation targets with their recorded roles. Seeds, prompt, and 10,000-token ceiling are shared; the contrast tests relation-guided integration within this budget.

\paragraph{Selection and generation.}
Task-keyword overlap selects seed objects with research-role coverage and a shared per-paper cap; internal edges join visible endpoints. Citation neighbors rank by keyword overlap in target metadata and role evidence, with paper-ID tie-breaking. The first two available records among the top ten are opened; absent neighbors leave seeds unchanged. Generation uses \texttt{qwen3.5-397b-a17b}, temperature 0, disabled thinking, and 2,800 output tokens. Proposals bind hypotheses and plans to visible object and dependency evidence.

The contrast evaluates the citation-guided expansion workflow as a whole: it changes access to neighboring evidence. The study does not isolate the contribution of relation labels or compare alternative neighbor-selection strategies.

\paragraph{Scientific quality.}
\texttt{MiniMax-M2.5} and \texttt{glm-5.1} independently score proposals against the same union of source excerpts, with condition labels and execution outcomes hidden. Reported means average both reviewers after the individual-rating transformation specified in Appendix~\ref{app:additional-results}. 

We report cross-paper integration directly and summarize the four scientific dimensions with an equal-weight descriptive composite:
\begin{equation}
 S_{\mathrm{sci}}=(N+G+I+Q)/4,
\end{equation}
where $N$ is novelty, $G$ evidence grounding, $I$ cross-paper integration, and $Q$ qualification handling. This descriptive composite and the independently judged Overall score both use 0--5. Table~\ref{tab:relation-expansion-detail} gives the component scores and family-balanced contrasts.

\begin{table*}[t]
\centering\small
\renewcommand{\arraystretch}{1.18}

\begin{tabular*}{\textwidth}{@{\extracolsep{\fill}}lrrrr@{}}
\toprule
Metric & Seed & Expanded & $\Delta$ & 95\% CI \\
\midrule
Cross-paper integration & \SPRAIntegrationSeed{} & \SPRAIntegrationExpanded{} & \SPRAIntegrationDelta{} & \SPRAIntegrationCI{} \\
Novelty & \SPRANoveltySeed{} & \SPRANoveltyExpanded{} & \SPRANoveltyDelta{} & \SPRANoveltyCI{} \\
Evidence grounding & \SPRAGroundingSeed{} & \SPRAGroundingExpanded{} & \SPRAGroundingDelta{} & \SPRAGroundingCI{} \\
Qualification & \SPRAQualificationSeed{} & \SPRAQualificationExpanded{} & \SPRAQualificationDelta{} & \SPRAQualificationCI{} \\
\bottomrule
\end{tabular*}
\caption{Scientific synthesis with SPRA relation expansion across all 1,001 paired questions; absolute scores use 0--5 and paired differences use score points. $\Delta$ is the equal-weight mean over 14 benchmark families; brackets give 95\% family-clustered bootstrap intervals. Seed and Expanded both use SPRA objects.}
\label{tab:relation-expansion-detail}
\end{table*}

\paragraph{Paired inference and evidence use.}
We average paired differences within each of 14 benchmark families, weight families equally, and bootstrap families 10,000 times. The full cohort includes the 28-question pilot and 973 remaining questions. The latter's integration contrast is \SPRAIntegrationHeldDelta{}/\SPRAExpansionScale{} (\SPRAIntegrationHeldCI{}), positive in all families; the composite is \SPRACompositeHeldDelta{} (\SPRACompositeHeldCI{}), positive in 13. Opened neighbors appear in \SPRAExpansionNeighborCount{}/1,001 expanded proposals.

\paragraph{Reproducibility materials.}
The accompanying evaluation package contains 4,004 numeric reviewer records, frozen citation edges, study protocol, and full/remaining-cohort estimates in \path{data/data/supplementary/paper_results.json}. The reported composites and intervals can be inspected alongside the underlying model-reviewer scores. The analysis includes all complete proposals and records reference conformance separately.

\FloatBarrier
\section{Research Objects in Use}
\label{app:use-cases}
\label{app:cases}

The three cases are drawn from the publicly available papers on speech-emotion unlearning \citep{ren2025unlearning}, personalized federated learning \citep{song2025fedaghn}, and multiscale network entropy \citep{brzovic2025entropy}. They illustrate how SPRA connects each paper's problem and contribution to its evaluation and findings, making the research argument easier to inspect and reuse.

\paragraph{Machine unlearning.}
The speech-emotion unlearning study of \citet{ren2025unlearning} asks how to remove sensitive information using only the forget set, without access to retained training examples. SPRA links this Problem to the proposed forget-set-only method, its unlearning-accuracy Measure, and the reported comparison with a baseline. An agent can retrieve the contribution and inspect the conditions attached to its result.

\paragraph{Personalized federated learning.}
In personalized federated learning, the FedAGHN case links static collaboration to the proposed method and reported overhead, scalability, and robustness findings \citep{song2025fedaghn}. These connected records let an agent inspect a method alongside its evaluation.

\paragraph{Multiscale network entropy.}
The multiscale-entropy case connects the limits of a single-scale description to a proposed framework and its evaluation using $R^2$, including a reported comparison of $0.92$ and $0.68$ \citep{brzovic2025entropy}.

\subsection{Application Interface}
\label{app:access}

\href{https://sciverse.opendatalab.com/}{Sciverse} demonstrates the SPRA interface through paper discovery, typed lookup, filtering, relation traversal, and source inspection. It presents research units and their relationships around a question, with links to source passages and tables. An offline walkthrough is included in the accompanying materials; RQ1--RQ3 evaluate scientific task performance.

\subsection{From a Research Question to Reusable Objects}
\label{app:public-quickstart}

Consider the speech-emotion unlearning example in Appendix~\ref{app:cases}. A researcher asks which proposed method removes information using only the forget set and under what evaluation conditions it succeeds. The access sequence is to discover the paper, select its Problem and Contribution, follow relations to the relevant Measure and Finding, and inspect the recorded source context. Keeping the paper and object identifiers with the selected evidence allows subsequent comparisons to refer to the same research objects.

\paragraph{REST access.} Create a Sciverse API Token through the \href{https://sciverse.opendatalab.com/}{Sciverse} account console and provide it as \texttt{SCIVERSE\_API\_TOKEN} in the local environment. Requests use a Bearer authorization header. The \href{https://sciverse.opendatalab.com/docs/sciverse/api/paper-schema}{Paper Schema API} accepts JSON request bodies for its POST operations. The examples use \url{https://api.sciverse.space} as the API base URL and \path{/paper-schema} as the service prefix. With the token configured, a paper search can be issued directly from a shell:

\noindent\begin{minipage}{\linewidth}
\begin{lstlisting}[basicstyle=\ttfamily\footnotesize,breaklines=true,columns=fullflexible]
BASE=https://api.sciverse.space
TOKEN="${SCIVERSE_API_TOKEN:?Set token}"
AUTH="Authorization: Bearer $TOKEN"
curl --fail-with-body \
  -X POST "$BASE/paper-schema/search" \
  -H "$AUTH" \
  -H "Content-Type: application/json" \
  -d '{"query":"speech emotion unlearning",
       "size":5}'
\end{lstlisting}
\end{minipage}

The response contains \texttt{items}, with \texttt{schema\_id} and paper metadata for each candidate, and \texttt{total}, which reports the match count and whether it is exact or a lower bound. Select the intended paper by its title and metadata. The following Python helper sends authenticated JSON requests:

\noindent\begin{minipage}{\linewidth}
\begin{lstlisting}[basicstyle=\ttfamily\footnotesize,breaklines=true,columns=fullflexible]
import json, os, urllib.request

base = "https://api.sciverse.space"
token = os.environ["SCIVERSE_API_TOKEN"]
headers = {
    "Authorization": "Bearer " + token,
    "Content-Type": "application/json"
}
def post(route, body):
    request = urllib.request.Request(
        base + "/paper-schema/" + route,
        data=json.dumps(body).encode(),
        headers=headers)
    with urllib.request.urlopen(
            request, timeout=60) as response:
        return json.load(response)

\end{lstlisting}
\end{minipage}

Enter the selected \texttt{schema\_id} from the search response to retrieve method-oriented materials:

\noindent\begin{minipage}{\linewidth}
\begin{lstlisting}[basicstyle=\ttfamily\footnotesize,breaklines=true,columns=fullflexible]
selected_id = input("schema_id: ").strip()
materials = post("materials", {
    "schema_ids": [selected_id],
    "goal": "method",
    "include_relation_context": True,
    "per_schema_entity_limit": 40,
    "per_schema_relation_limit": 40,
    "per_schema_attribute_limit": 40
})
\end{lstlisting}
\end{minipage}

Each returned item contains \texttt{schema\_paper}, \texttt{entities}, \texttt{relations}, \texttt{evidence}, and \texttt{terms}. Per-resource counts expose \texttt{total}, \texttt{returned}, and \texttt{truncated}. A material package selects content for the requested goal; a complete object listing requires pagination through the corresponding entity or relation interfaces. The capability endpoint lists the current service types; callers should use these names when selecting resources.

For an object or relation with source markers, \path{/paper-schema/resolve-provenance} accepts its \texttt{schema\_id} and returned \texttt{marker\_nums}, together with a bounded context window. Preserve these returned locators; do not derive them from section numbers. The response provides source segments and unresolved locators. Inspect warnings when \texttt{partial} is true and retrieve additional pages or narrower selections when \texttt{truncated} is true. These examples specify the access procedure; the paper's quantitative findings come from the frozen evaluations.

\paragraph{Capabilities and errors.} An authenticated \texttt{GET} to \path{/paper-schema} returns the current contract version, supported resource types, and limits. HTTP 401 indicates an authentication problem; HTTP 403 indicates insufficient account permissions. For HTTP 429, clients should wait according to \texttt{Retry-After} or the response's \texttt{retry\_after} value. The returned \texttt{X-Request-ID} supports request tracing.

\paragraph{Agent access.} The \href{https://sciverse.opendatalab.com/docs/sciverse/skills/paper-schema}{Paper Schema Agent Skill documentation} provides installation and configuration instructions. An agent can use \path{search_paper_schemas}, \path{query_paper_entities}, \path{query_paper_relations}, and \path{resolve_paper_context} for the same sequence, or \path{build_paper_materials} for task-specific selection. Both interfaces retain paper-scoped identities. Related-paper discovery and resolved citation traversal support expansion to other papers; names shared across papers do not by themselves establish object identity.

\section{Reproducibility and Resource Use}
\label{app:artifact}
\label{app:reproducibility}

\subsection{Evaluation Materials and Reanalysis}

The supplements provide code, tests, model settings, source locators, corpus statistics, task-level results, SPRA examples, and frozen inputs for offline inspection and reanalysis under the terms in Appendix~\ref{app:resource-access}.

\path{software.zip} contains extraction code, prompts, schemas, analysis and replay scripts, and tests. \path{data.zip} contains study records and fixtures plus the complete 9,849-paper aligned RQ1 cohort, with complete-text, retrieved-passage, and SPRA inputs for each item. The archive also records the 9,996-paper candidate catalog and 147 exclusion records: 12 candidates lack a positive Text-RAG candidate, and 135 are removed to balance domain--task cells. Each eligible paper has one saved three-representation input set; study-specific complete-case counts remain as reported in Appendix~\ref{app:additional-results}. Matched inputs support offline inspection; saved study outputs support numerical reanalysis without API access. The archives create sibling \path{software/} and \path{data/} directories and include file checksums. The aligned inputs are in \path{data/data/main/}; supporting records are in \path{data/data/supplementary/}.

\noindent\begin{minipage}{\linewidth}
\paragraph{Offline reproduction.}
Extract both archives into the same parent directory. From their shared parent directory, run the packaged consistency check and regression tests:

\begin{lstlisting}[basicstyle=\ttfamily\footnotesize,breaklines=true,columns=fullflexible]
python3 software/software/verify_release.py
python3 -m unittest discover \
  -s software/code
\end{lstlisting}
\end{minipage}

The complete analysis commands are documented in \path{software/REPRODUCE.md}. Numerical reanalysis uses saved records without hosted models; fresh extraction or answer generation requires source articles and hosted services.

\subsection{Executable Implementation and Tests}
\label{app:software}

The \href{https://github.com/gavinwxy/ai-paper-structured-extraction}{SPRA extraction implementation} is included with the evaluation code. Its prompts and JSON schemas define the extraction stages; assembly code validates identifiers, relation endpoints, and source bindings. Appendix~\ref{app:construction-spec} describes the construction procedure.

The regression tests check selected reported values and their arithmetic against fixtures in the data archive. \path{software/REPRODUCE.md} documents the offline commands and the source articles and hosted services required for fresh model runs.

\subsection{Model Configurations and Request Settings}
\label{app:model-call-snapshot}

Llama 4 Scout is the primary Reader for RQ1--RQ3; DeepSeek-v4-pro generates source-grounded RQ1/RQ2 references and audits the RQ3 implementation; MiniMax-M3 scores RQ1/RQ2 answers against shared references; and MiniMax-M2.5, Qwen3.5-27B, and GLM-5.1 independently review RQ3 proposals. The supplementary \path{data/data/supplementary/model_settings.json} records model aliases, role-level output ceilings, and backend-specific request options. Primary evaluation calls use temperature 0; these per-call settings are distinct from Reader input-context budgets.

\paragraph{Construction configuration.} The reference extractor uses \texttt{qwen3.5-35b-a3b}, temperature 0, disabled thinking, and output ceilings of 32,768 tokens for content and 24,576 for planning auxiliaries. These settings describe the reference extractor; the corpus statistics summarize the deployed collection rather than a single-model construction run.

\paragraph{Evaluation request limits.} The frozen registry records temperature 0 and top-$p$ 1 for RQ1--RQ3. RQ1 output ceilings are 24,000 tokens for the Reader, 8,000 for references, and 2,400 for judging. RQ2 uses 3,500/6,000 for directed/Top-5 answers, 12,000 for references, and 2,400 for judging. RQ3 uses 7,000 for digestion, 6,500 for candidate generation, 1,800 for binding, 2,200 for implementation assessment, and 2,400 for review. These are configured output limits; actual consumption is reported separately. Hosted APIs manage model hardware, so the recorded aliases and request settings identify the evaluated services.

\subsection{Task-Based Assessment of Extracted Content}
\label{app:extraction-use}
\label{app:finding-reuse}

RQ1 assesses the functional adequacy of the extracted representation through source-conditioned scientific QA and structured operations. DeepSeek-v4-pro generates shared answer references from Full Compact, the complete parsed-text input; MiniMax-M3, blind to condition, scores Reader responses against them. G-U1 measures correctness, coverage, and qualification against these references; G-U3 focuses on Problem + Contribution. Source traceability and complete relation endpoints support inspection and research-object operations. The supplementary materials also document the sampled object assessment.

\subsection{Token-Accounting Scope}
\label{app:cost}

Storage, retrieval, and maintenance are supplied by Sciverse. Token totals measure complete inputs and outputs for the retained task calls. They quantify the evaluated workflows' text volume; they do not represent the service bill or include storage and retrieval costs. Appendix~\ref{app:additional-results} specifies the included stages and excluded calls for each study.

\FloatBarrier

\end{document}